\documentclass[journal]{IEEEtran}

\ifCLASSINFOpdf
\else
\fi

\usepackage{svg}
\usepackage{cite}
\usepackage{hyperref}
\usepackage{booktabs}
\usepackage{multirow}
\usepackage{amsmath}
\usepackage{algorithmicx}
\usepackage[ruled,linesnumbered,nofillcomment]{algorithm2e}

\begin{document}
%
\title{A Fast and Robust Navigation System for the High-speed Flight of UAVs}
\title{HEPP: Hyper-efficient Perception and Planning for High-speed Obstacle Avoidance of UAVs}
%
%
%

\author{Anonymous Authors
\thanks{Authors' affiliations and funding acknowledgements. }
}

\author{Minghao Lu$^1$, 
        Xiyu Fan$^1$,
       Bowen Xu$^1$,
       Zexuan Yan$^1$,
       Rui Peng$^1$,
       Han Chen$^2$, 
       Lixian Zhang$^{3*}$,
       and Peng Lu$^{1*}$
\thanks{$^1$Minghao Lu, Xiyu Fan, Bowen Xu, Zexuan Yan, Rui Peng and Peng Lu are with the Adaptive Robotic Controls Lab (ArcLab), Department of Mechanical Engineering, The University of Hong Kong, Hong Kong, SAR, China (e-mail: minghao0@connect.hku.hk; fanxiyu@connect.hku.hk; link.bowenxu.connect.hku.hk;  ryan2002@connect.hku.hk; pengrui-rio@connect.hku.hk; lupeng@hku.hk). $^2$Han Chen is with Northwestern Polytechnical University, Xi’an, Shanxi, China (e-mail: stark.chen@connect.polyu.hk). $^3$Lixian Zhang is with Harbin Institute of Technology, Harbin, Heilongjiang, China (e-mail: lixianzhang@hit.edu.cn). $^*$Corresponding author.}
\thanks{
       This work was supported by General Research Fund under Grant 17204222, and in part by the Seed Fund for Collaborative Research and General Funding Scheme-HKU-TCL Joint Research Center for Artificial Intelligence.
}%
}

\maketitle

\begin{abstract}
High-speed obstacle avoidance of uncrewed aerial vehicles (UAVs) in cluttered environments is a significant challenge. Existing UAV planning and obstacle avoidance systems can only fly at moderate speeds or at high speeds over empty or sparse fields. In this article, we propose a hyper-efficient perception and planning system for the high-speed obstacle avoidance of UAVs. The system mainly consists of three modules: 1) A novel incremental robocentric mapping method with distance and gradient information, which takes 89.5\% less time compared to existing methods. 2) A novel obstacle-aware topological path search method that generates multiple distinct paths. 3) An adaptive gradient-based high-speed trajectory generation method with a novel time pre-allocation algorithm. With these innovations, the system has an excellent real-time performance with only milliseconds latency in each iteration, taking 79.24\% less time than existing methods at high speeds (15 m/s in cluttered environments), allowing UAVs to fly swiftly and avoid obstacles in cluttered environments. The planned trajectory of the UAV is close to the global optimum in both temporal and spatial domains. 
Finally, extensive validations in both simulation and real-world experiments demonstrate the effectiveness of our proposed system for high-speed navigation in cluttered environments.
\end{abstract}


%
\IEEEpeerreviewmaketitle

\section{Introduction}
%
%
%
%
\IEEEPARstart{W}{ith} the development of robotic autonomous technology, uncrewed aerial vehicles (UAVs) have become pivotal in various applications, ranging from surveillance to search and rescue missions \cite{mohsan2022towards}. Ensuring the safe and efficient navigation of UAVs is paramount for their successful deployment. Existing works have shown a robust performance in flying at a moderate speed \cite{marshall2021survey}.


Nevertheless, high-speed obstacle avoidance in unknown environments remains challenging and potentially hazardous. In high-speed flight scenarios, UAVs face unique challenges related to the perception accuracy and response time of the system. Typically, an autonomous obstacle avoidance system has two main components: mapping and motion planning. The function of the mapping module is to model the environment using sensor data. The limited field of view (FOV) and noise of RGBD cameras make it difficult to ensure flight safety, while Lidar sensors boast high-accuracy measurement, the extensive computation of point clouds poses challenges to real-time performance. The motion planning module generates feasible actions based on perception. When the environment is extremely cluttered and complex, it is difficult to find feasible solutions at the desired high speed. Furthermore, replanning under high-speed motion must be completed in an extremely short duration. All of these issues raise the risk of crashes when UAVs fly at high speeds in cluttered environments.

\begin{figure}[t]
\centerline{\includegraphics[scale=0.51]{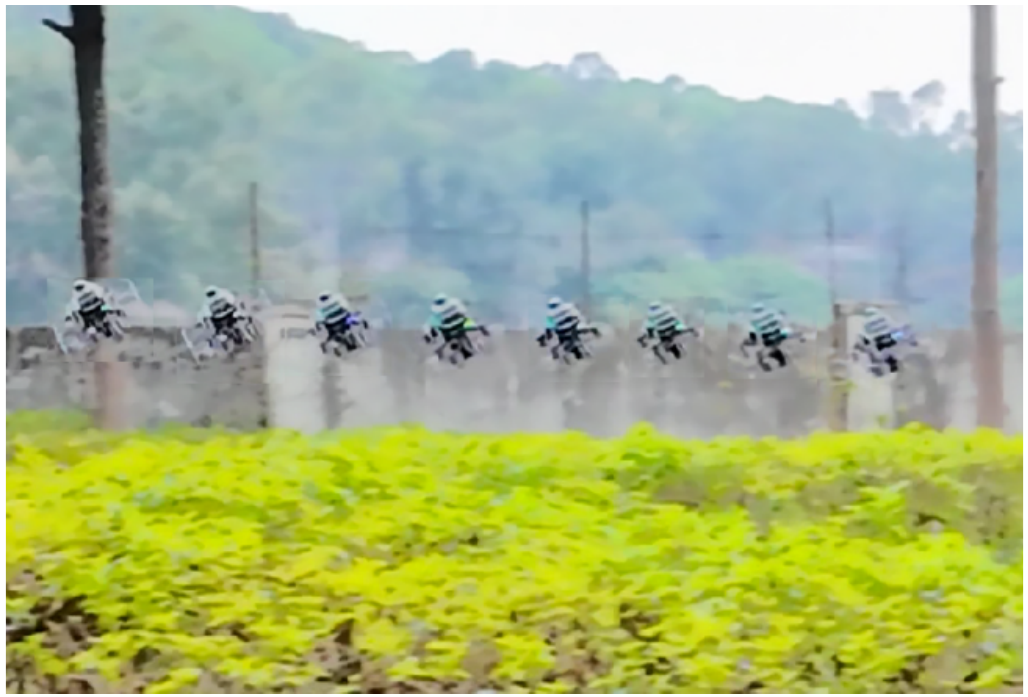}}
\caption{A demonstration of high-speed navigation, the UAV flies at over 11.0 m/s across the forest.}
\label{f: cover}
\end{figure}

\begin{figure}[h]
\centerline{\includegraphics[scale=0.38]{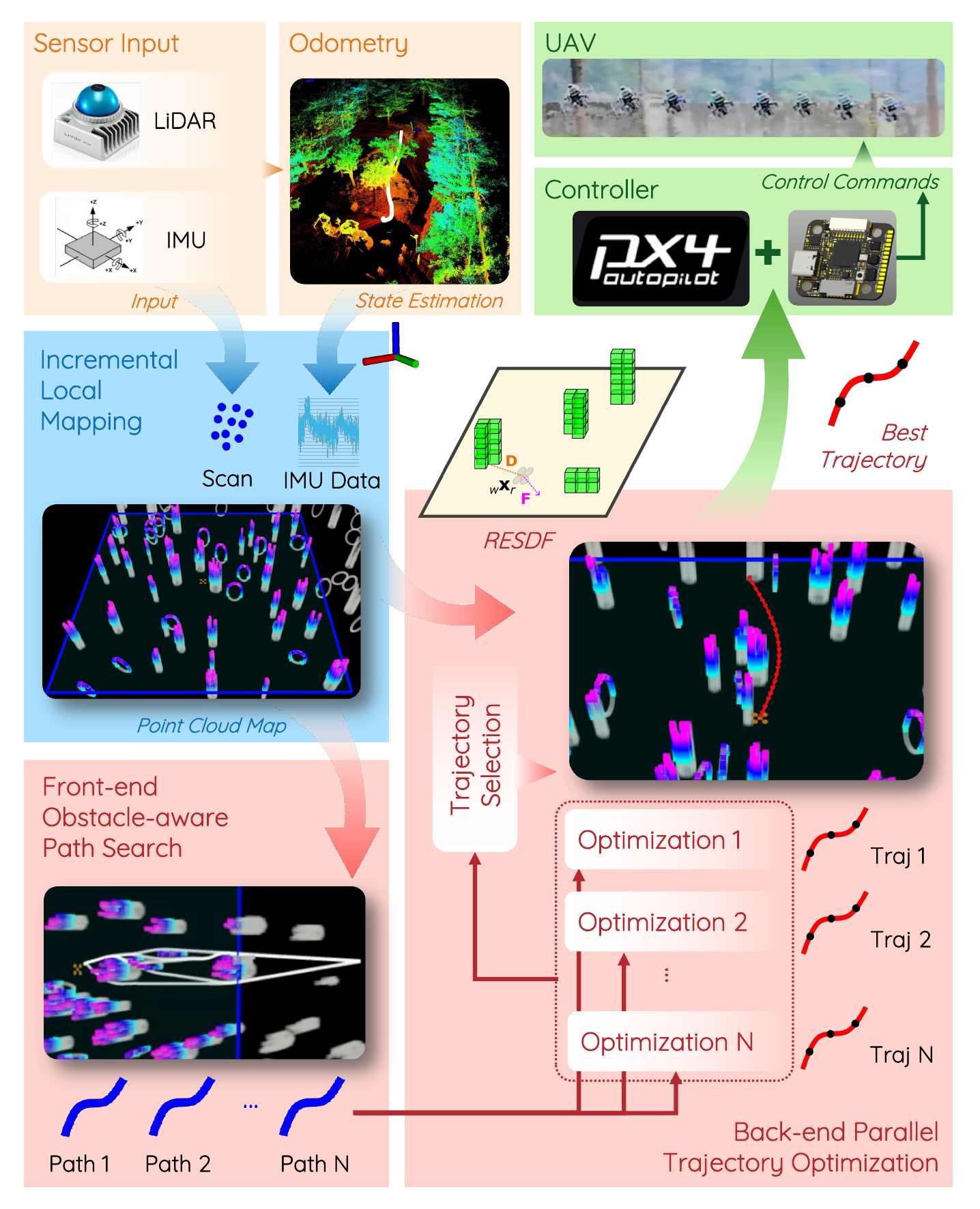}}
\caption{An overview of our system framework.}
\label{f: frame}
\end{figure}

In this article, we propose \textbf{HEPP}: \textbf{H}yper-\textbf{E}fficient \textbf{P}erception and \textbf{P}lanning for high-speed obstacle avoidance of UAVs, which aggregates all the mentioned difficulties in perception and planning, as shown in Fig. \ref{f: cover}. The overview of our proposed system is shown in Fig. \ref{f: frame}. The system utilizes a high-accuracy Lidar and an IMU sensor as raw data input to estimate the environment model and the UAV states, and then plan the UAV motion by a front-end topological path search and a back-end trajectory generation. For each path guiding trajectory optimization, we approximate the global optimum by selecting the one with the minimum cost. 
The main contributions of this article are summarized as follows:
\begin{itemize}
\item[1)] We propose a novel mapping method with a new data structure. A local map can be maintained and updated with very high time and memory efficiency, and distance and gradient information can be obtained directly from the map without additional computations, which is suitable for high-resolution sensors and large-scale scenarios. 

\item[2)] We propose a novel obstacle-aware topological path search algorithm. The algorithm can generate multiple non-redundant paths in complex cluttered environments with very low time cost, which facilitates the approximation towards global optimality of UAV motion planning.

\item[3)] We propose an adaptive gradient-based high-speed trajectory planning method. With the adaptive trajectory initialization, the planner can find a high-quality trajectory by numerical optimization that satisfies safety and dynamic feasibility. By selecting the best trajectory from the multiple trajectories, the UAV motion is further closer to the global optimum.

\item[4)] We propose a fully autonomous UAV system (including mapping, front-end path search, and trajectory planning) for high-speed navigation. We validate the performance of our entire system in various simulation and experimental tests. The latency of the entire system is within a few milliseconds, which is highly efficient.

\end{itemize}

\section{Related Works}

\subsection{Local Mapping and ESDF} 
In most robotic applications, the occupancy grid map is used for navigation and obstacle collision checks due to its intuitive principle and simple implementation \cite{mapreview}. The process usually consists of raycasting, updating the grid state, and so on. Based on this, the Euclidean Signed Distance Field (ESDF) \cite{esdf2016} is widely used in gradient-based robotic trajectory optimization, where the distance and gradient information to obstacles are necessary, such as \cite{raptor, tao2023seer, pankert2020perceptive}. The generation of ESDF is based on the map's data structure. In \cite{fiesta, vblox, voxfield, fiimap}, the environment is formulated by a fixed-size 3D map divided by uniform grids. Firstly, raycasting is used to update the probability value of the grids, where the state of each grid will be classified as occupied, free, or unknown. Secondly, obstacle inflation is applied to the occupied space to guarantee safety. Finally, ESDF is calculated by Breadth First Search (BFS) and trilinear interpolation. However, these uniform grid-based methods have extensive memory consumption, making them impractical for long-distance flight. Moreover, modules such as obstacle inflation, especially ESDF, will bring a very high computational load and cannot run in real-time on high-resolution Lidar sensors. Robocentric mapping methods \cite{rogmap, universal} only maintain the robocentric local map using a zero-copy map sliding strategy, significantly reducing memory usage. However, these methods still have the limitation of updating ESDF in real-time. Compared to uniform grid-based data structure, octree-based mapping methods \cite{octomap, ufomap, multires} are more memory efficient. The calculation of the ESDF relies on the Nearest Neighbor Search (NNS) of octree, which is also more convenient and efficient to use. However, these methods also use the fixed map origin, which will cause extensive memory usage, and re-building octree each frame is not possible in real-time. \cite{ioctree} proposes an incremental octree data structure for the point cloud, which is time and memory-efficient for Lidar point cloud mapping. However, this work does not present a complete methodology of mapping. Inspired by the above methods, we will propose an incremental robocentric point cloud mapping method based on \cite{ioctree} to address all the problems mentioned.

\subsection{Topological Path Searching} 
Topological path searching aims to find paths belonging to different homology classes \cite{jaillet2008path, bhattacharya2010search, bhattacharya2012topological}. A graph-based representation called roadmap is used to approximate the continuous configuration space. Probabilistic Roadmap (PRM) \cite{prm1996} is the most widely used roadmap type. \cite{ctop,raptor,zhang2019online, jin2023improved} are the current SOTA works based on PRM. The generation of PRM is firstly sampling in the environment to get a series of nodes and then connecting the adjacent nodes that satisfy desired principles such as visible condition \cite{simeon2000visibility}. After PRM is constructed, any search algorithm, such as Dual-Dijkstra, is used in the graph where paths with different topologies can be found, but many of the output paths still belong to the same homology class. Therefore, the phase then proceeds to check and prune the redundant paths is necessary. PRM-based methods are easy to implement and can capture relatively complete topology classes. However, the performance of these methods is limited by the sampling strategy and resolution in the space. A dense sampling will cause extremely time-consuming graph construction and path searching, while a sparse sampling will lead to lower quality paths or even failed path searches. In addition, the path pruning is also an exhaustive process, especially in dense graphs with a large number of nodes. Another type of roadmap construction approach is based on environmental geometry. \cite{yang2022far, oapr} represents the roadmap by formulating the obstacles as polygons, and \cite{song2024fht} uses polygon corridors to represent the free space.
Based on the comprehensive environmental geometry information, these methods eliminate the path pruning process and can capture distinct paths without being redundant directly. However, the point cluster and convex hull generation are still very time consuming when the point cloud is dense, and they are difficult to adapt to scenarios containing complex and nonconvex obstacles such as circles and narrow gaps. \cite{egov2} captures the distinct paths by searching on opposite sides of the obstacles, which is more time-efficient compared to the previous works. However, the method can only generate very few paths and is not suitable for complex 3D environments. There are also some semantic graph-based topological planning methods, such as \cite{kim2023topological, cailhol2019multi, an2024etpnav}. These works can be utilized for specialized navigation tasks, however, learning-based approaches struggle to work robustly in general environments and typically necessitate GPU acceleration for real-time performance. These methods cannot satisfy the demands of high-speed flight due to the limitations of robustness and computational load. 

\subsection{High-speed Trajectory Planning of UAVs} 
Various approaches have been proposed to enable high-speed autonomous quadrotor trajectory generation. Existing methods for UAVs' trajectory planning can be divided into four main streams: polynomial-based methods, Bézier curves or B-Splines-based methods, $\mathbf{MINCO}$-based methods, and learning-based methods. Due to the intuitive parameterization and efficiency, the polynomial trajectory is widely used in UAV applications. \cite{minisnap} proposes a minimum snap polynomial trajectory with waypoints and dynamics constraints, which achieves quadrotor flying through the hoops with a speed of up to 2.6 m/s. Based on \cite{minisnap}, \cite{sfc} extends the work to real-time navigation in unknown environments by enforcing corridor constraints. \cite{agile2024saska} addresses the real-time planning of minimum-time trajectories over multiple waypoints based on a time-allocation algorithm. The UAV can fly at 100 km/h in a free space. As for B-spline or Bézier curve-based methods, \cite{usenko2017real, raptor} define the trajectory optimization problems based on uniform B-splines, and \cite{faster, sahingoz2014generation} use Bézier curves as the trajectory formulation. These works can make UAVs navigate in unknown environments at a speed under 5 m/s with a robust performance. However, in order to satisfy the continuity of the high-order states (e.g, position, velocity, acceleration) of the trajectory, polynomials and Bézier curves need to introduce additional equality constraints. Besides, the control points of Bézier curves and B-Splines are not on the trajectory, which is not intuitive. In terms of temporal optimization, all these methods are temporal-spatially coupled. The isolated optimization of the temporal domain may result in uncontrollable spatial properties of the trajectory, such as safety and feasibility. These limitations confine these methods to navigating at relatively conservative speeds in unknown environments or only flying at high speeds over an open terrain. To address the problems, \cite{gcopter} proposes a trajectory class $\mathbf{MINCO}$ that retains local smoothness by spatial-temporal deformation via linear-complexity operations. \cite{egov2, fastdodge, lu2024fapp} utilize $\mathbf{MINCO}$ and complete more complex tasks, including dynamic obstacle avoidance and swarm formation. However, the $\mathbf{MINCO}$-based unconstrained optimization problems rely on a high-quality initial value. When the environment is cluttered and the speed is high, the optimization problem will be highly non-convex. These methods struggle to find a solution in these cases. By a receding corridor strategy, \cite{bubble} achieves flight speeds over 13 m/s in real-time. However, this method still does not address the issues of initial values and global optimality, thus restricting it to very sparse environments. Learning-based methods have recently gained significant popularity as well. \cite{learningsr, song2023rl, zhang2024back} uses network policy to generate a trajectory directly from the depth image and current state. Limited by the sensor type and the training environments, the method lacks robustness, and the flight speed cannot exceed the speed at which it was trained. We will devise a novel approach enabling UAVs to plan safe and smooth high-speed trajectories in generic dense environments.

\section{Incremental Robocentric Point Cloud Map}
In this section, we will present our proposed incremental robocentric point cloud map (IROP-Map) algorithm. The point cloud-based map maintains a local map moving with the robot and releases the free space by a novel efficient raycasting method, making it suitable to operate at high resolution and in large-scale environments. A novel Robocentric ESDF (RESDF) is proposed to evaluate the obstacle avoidance cost. Compared to the traditional ESDF methods \cite{fiesta, vblox, voxfield, fiimap} that require iterative updating, our method can directly obtain the distance and gradient at a random position, which significantly reduces the computational load. Furthermore, the obstacle inflation process is eliminated compared with the occupancy grid map, which also decreases the mapping time cost.


\subsection{Map Structure} 
Assuming at the $k$th time stamp $t_k$, a mobile robot is at position $_{w}\mathbf{p}_k \in \mathbf{R}^3$ in the world frame $w$. We define the local map as a cubic space $S_k$ with center point $_{w}\mathbf{p}_k$ and size $\mathbf{s} = (s_x, s_y, s_z) \in \mathbf{R}^3$. The local map maintains the point set $_{w}M_k \subseteq S_k$ at time $t_k$ by an $i$-$Octree$ \cite{ioctree} with a minimal extent $e_{min}$. Using $i$-$Octree$, we can dynamically add or delete points on the map. Besides, fast Nearest Neighbor Search (NNS) such as $K$-nearest neighbors search $\mathcal{K}$ and radius neighbors search $\mathcal{R}$ can be easily applied in the following sections. The dynamic update process will be demonstrated in the following part.

\subsection{Incremental Update}
\label{s:3B}
\begin{figure}[t]
\centerline{\includegraphics[scale=0.52]{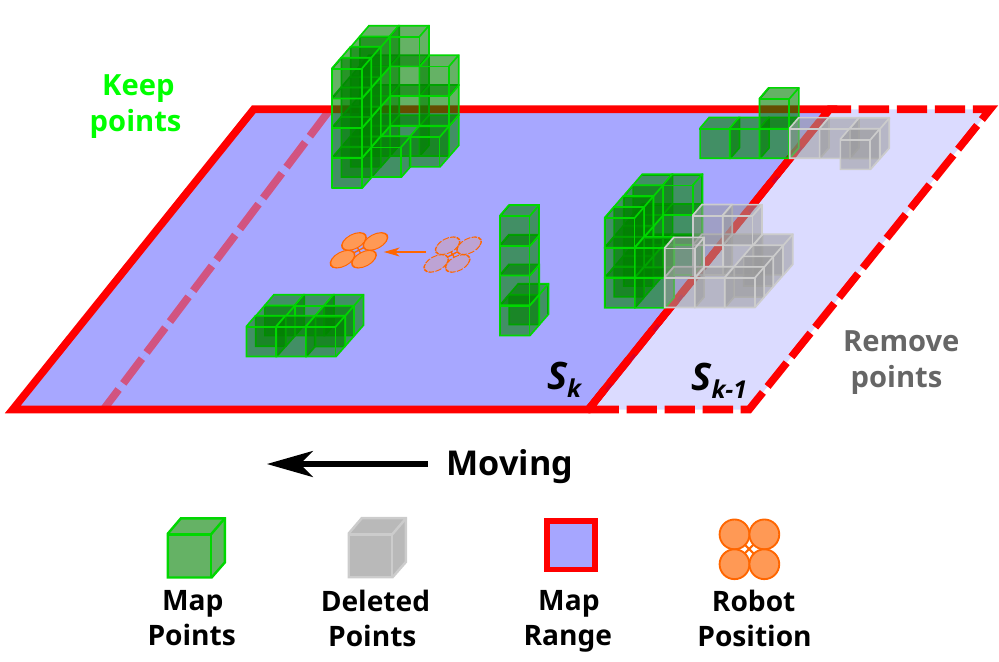}}
\caption{The illustration of the map sliding process. The points in the old map range that do not intersect with the current map range are deleted to release the memory usage.}
\label{f: map_moving}
\end{figure}

At time $t_k$, the sensor scans a frame of point cloud $_{w}C_k$ in the local map range $S_k$. Firstly, we filter the points to make the points distributed uniformly. For each arbitrary 3D point $_{w}\mathbf{o} = (_{w}o_x, _{w}o_y, _{w}o_z)$ in $_{w}C_k$, we perform a transformation $f: {_w}\mathbf{o} \rightarrow {_w}\mathbf{o}'$ as:
\begin{equation}
f(_{w}\mathbf{o}) = (\lfloor _{w}o_{x,y,z}/r \rfloor + 0.5)r ,
\end{equation}
where $\mathbf{o}'$ is the point after transformation and $r$ represents the minimum distance between every two points in each axis direction. Then, we obtain the filtered point cloud $_{w}C'_k$ as the input of our map updating algorithm, which comprises three steps:

\emph{1) Map sliding}:
As the robot moves from $_{w}\mathbf{p}_{k-1}$ to $_{w}\mathbf{p}_k$, the range of the local map changes from $S_{k-1}$ to $S_k$, as shown in Fig. \ref{f: map_moving}. In the first step of the update, we remove the points in the old map $_{w}M_{k-1} \subseteq S_{k-1}$ that do not intersect with the new map space $S_k$, so this process can be expressed as:
\begin{equation}
_{w}M^{1}_{k} = {_w}M_{k-1} \cap S_k ,
\end{equation}
where $_{w}M^{1}_{k}$ denotes the one-step updated point set at $t_k$. The points that need to be removed can be searched via $i$-$Octree$ and deleted. If a leaf node of the tree contains no points after deletion, it will be deleted as well to release the memory space. In contrast, the grid-based mapping methods allocate memory for all grids in the map and need a data copy process in the memory while updating, which is not suitable for large-scale scenarios. 

\emph{2) Raycasting}: In most existing works, the raycasting process needs to update each grid on the occupancy grid map that the rays pass through, which is very time-consuming. We propose an efficient raycasting method based on our map structure.  Firstly, we perform an $SE(3)$ transformation $_{b}^{w}T_{k}$ from the global frame $w$ to the local sensor frame $b$, and obtain $_{b}C_k$ and $_{b}M^{1}_{k}$. $_{b}^{w}T_{k}$ can be obtained by using the LIO algorithm such as \cite{fastlio}. Then, we project the current scan points into a $m\times n$ depth image $I^C_k \in \mathbf{R}^{m\times n}$, $m$ and $n$ are determined by the angle resolution  $\delta$ of the polar coordinate space. Thus, for an arbitrary point $_{b}o = (_{b}o_x, _{b}o_y, _{b}o_z)$ in $_{b}C_k$, the projected pixel index $(i,j)$ can be determined by:
\begin{equation}
i = \lfloor atan(\frac{_{b}o_y}{_{b}o_x})/\delta \rfloor, \ j = \lfloor atan(\frac{_{b}o_z}{\sqrt{_{b}o^2_x + _{b}o^2_y)}})/\delta \rfloor,
\end{equation}
and then we can calculate the value of the pixel $(i,j)$ as:
\begin{equation}
I^C_k(i,j) = \min\limits_{_{b}\mathbf{o} \in _{b}C_{k,ij}} \Vert _{b}\mathbf{o} \Vert,
\end{equation}
where $_{b}C_{k,ij}$ are the overall points that projected on the pixel $(i,j)$. 

\begin{figure}[t]
\centerline{\includegraphics[scale=0.27]{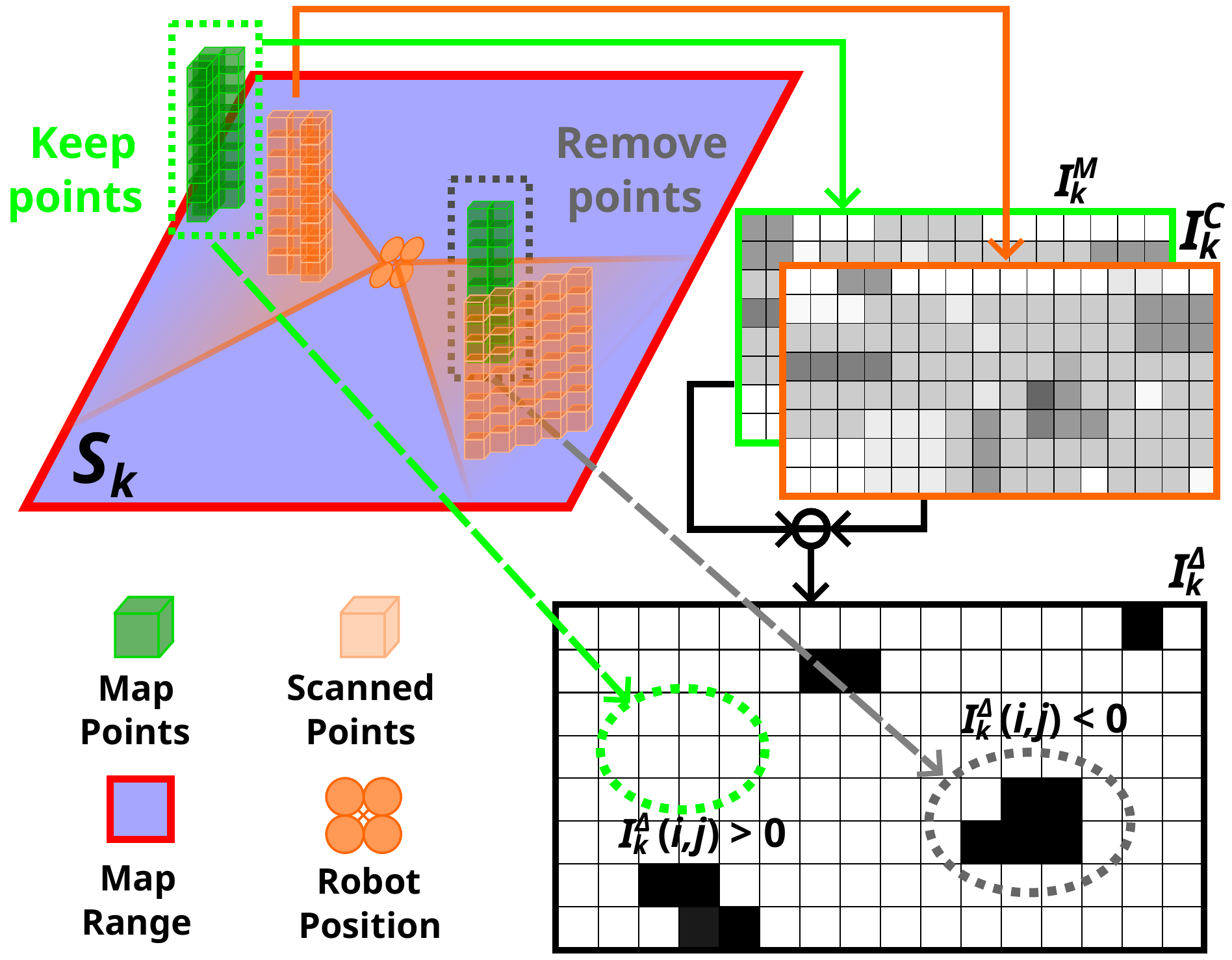}}
\caption{The illustration of the raycasting process. The point will be removed if it is passed by a new laser scan, where the noise and the dynamic points can be eliminated.}
\label{f: raycasting}
\end{figure}

In the same way, the projected depth image $I^M_k$ of $_{b}M^{1}_{k}$ can be acquired. Hitherto, the raycasting process can be formulated as a matrix element-wise subtraction between $I^M_k$ and $I^C_k$:
\begin{equation}
I^{\Delta}_k = I^M_k - I^C_k,
\end{equation}
where $I^{\Delta}_k$ is the difference image. If a pixel value satisfies $I^{\Delta}_k(i,j) < 0$, it means that the previous map points $_{b}M^1_{k,ij}$ projected on $(i,j)$ are passed by a new shot ray, and the corresponding points in the world frame $_{w}M^1_{k,ij}$ will be deleted in the tree in the same way as in \emph{1)}. The map points after raycasting $_{w}M^{2}_{k}$ can be expressed as:
\begin{equation}
_{w}M^{2}_{k} = \{_{w}M^1_{k,ij} | I^{\Delta}_k(i,j) > 0\},
\end{equation}
The raycasting process is illustrated in Fig. \ref{f: raycasting}.

\emph{3) Adding new points}: Finally, we incrementally insert new points to the tree based on the latest filtered points $_{w}C'_k$. If there are points out of the previous range of $i$-$Octree$, the nodes of the tree will be expanded incrementally. Besides, to decrease unnecessary points and make points query more efficient, we down-sample the input points $_{w}C'_k$ while inserting. If the extent of a node is less than $e_{min}$ and the number of points within this node is larger than $n_{max}$, points will not be added to this node. We define the down-sampled input points as $_{w}C''_k$. Consequently, the final output of the updated local map points $_{w}M_k$ at time $t_k$ can be expressed as:
\begin{equation}
_{w}M_{k} = {_w}M^2_{k} \cup {_w}C''_k ,
\end{equation}

\subsection{Occupancy Check}
Using $i$-$Octree$, the occupancy check at a 3D position $_{w}\mathbf{x} \in \mathbf{R}^3$ can be performed based on the radius neighbors search function $\mathcal{R}$. As mentioned in subsection~\ref{s:3B}, the minimum distance between every two points in each axis is $r$. Thus, we search for map points at $_{w}\mathbf{x}$ in radius $r$, and the state of position $_{w}\mathbf{x}$ can be described by:
\begin{equation}
\mathrm{state} (_{w}\mathbf{x}) = \{
\begin{array}{ll}
\mathrm{Occupied}, & \mathcal{R}(_w\mathbf{x}, r) \neq \emptyset \\
\mathrm{Free}, & \mathcal{R}(_w\mathbf{x}, r) = \emptyset
\end{array},
\end{equation}
If no points are searched in radius $r$, $_{w}\mathbf{x}$ is considered to be free, otherwise is occupied.

\subsection{RESDF}
\label{s:3D}
\begin{figure}[h]
\centerline{\includegraphics[scale=0.25]{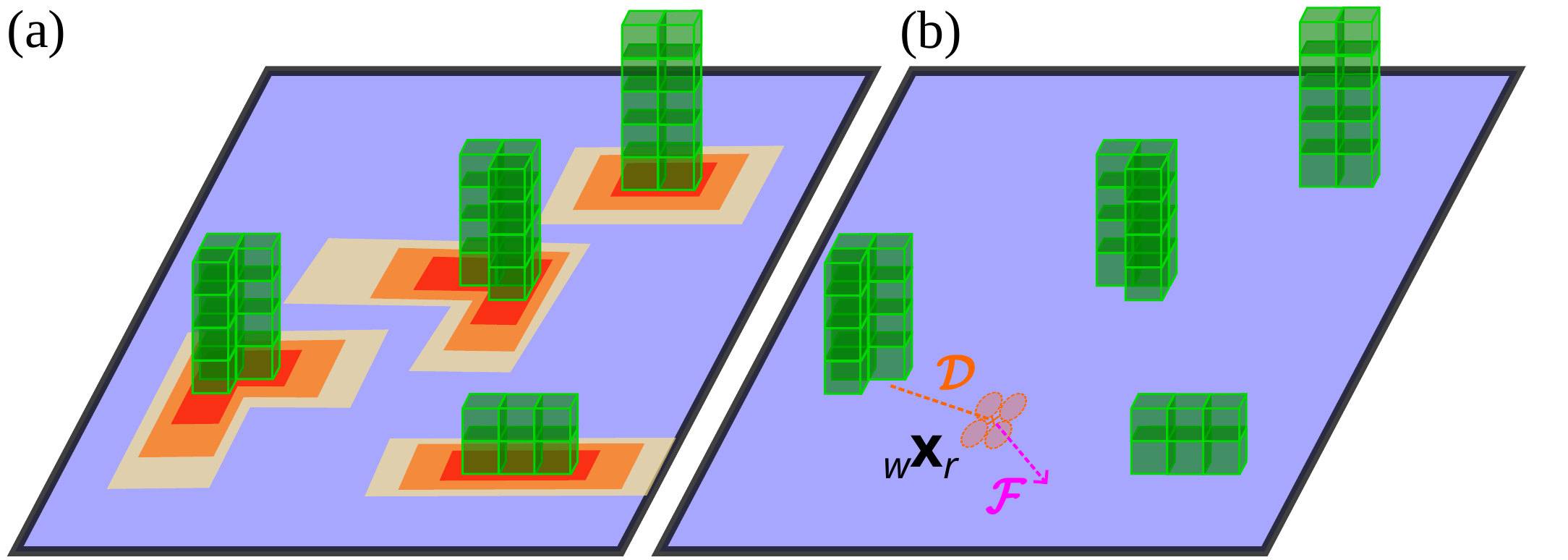}}
\caption{The comparison of traditional ESDF (a) and our RESDF (b). ESDF needs to be updated in the whole map, and our method can obtain the distance and gradient directly.}
\label{f: esdf}
\end{figure}

Here, we present our Robocentric Euclidean Signed Distance Field (RESDF) method. 
The RESDF is constructed only in the adjacent space of the robot center position ${_w}\mathbf{x}_r \in \mathbf{R}^3$. 
It is constructed by the $K$-nearest neighbors search function $\mathcal{K}$ of our proposed map. 
With $\mathcal{K}({_w}\mathbf{x} ,1)$, we can acquire the nearest neighbor point to position ${_w}\mathbf{x}$ in the map. In the wake of this, the RESDF at ${_w}\mathbf{x}_r$ can be calculated by:
\begin{equation}
\mathcal{D} ({_w}\mathbf{x}_r) = ||\mathcal{K}({_w}\mathbf{x}_r ,1) - {_w}\mathbf{x}_r||,
\label{eq: dist}
\end{equation}
\begin{equation}
\mathcal{F}_{\{x,y,z\}} ({_w}\mathbf{x}_r) = \frac{\mathcal{D} ({_w}\mathbf{x}_r) - \mathcal{D} ({_w}\mathbf{x}_r + \epsilon_{\{x,y,z\}})}{ ||\epsilon_{\{x,y,z\}}|| },
\label{eq: dist_grad}
\end{equation}
where $\mathcal{D}$ is the distance value function, $\mathcal{D}({_w}\mathbf{x}_r)$ indicates the distance value of the robot from the nearest obstacle. $\mathcal{F} = -\nabla \mathcal{D}$ is the negative gradient of distance field. $\epsilon_{\{x,y,z\}} \in \mathbf{R}^3$ is a tiny positive offset along each axis ${x,y,z}$. The gradient value at ${_w}\mathbf{x}_r$ is approximated by the difference of the distance field on each axis. The elaboration of RESDF construction is shown in Fig. \ref{f: esdf}.

Compared to conventional ESDF methods \cite{fiesta, vblox, voxfield, fiimap}, our method does not require real-time updates based on the entire environment information obtained from sensors. The distance and gradient value can be calculated directly at an arbitrary position owing to our novel map structure, which significantly reduces the computational overhead and can be easily used in gradient-based optimizations.

\section{Obstacle-aware Topological Path Searching}
In this section, we propose our obstacle-aware topological path-searching algorithm. Compared to most used methods like probabilistic roadmap (PRM) \cite{raptor} and geometry-based topological planning methods \cite{oapr,egov2}, our algorithm can search a group of distinct paths covering the space thoroughly and significantly reduces the computation time. The distinct paths avoid motion planning to fall into a local minimum, which is vital for high-speed flight.


\subsection{Obstacle-aware Topology Graph Structure}
For a mobile robot in navigation with a start position $_{w}\mathbf{p}_{start} \in \mathbf{R}^3$ and target position $_{w}\mathbf{p}_{end} \in \mathbf{R}^3$, we expect to find multiple non-homogeneous collision-free paths. In this article, we define the topological path $\tau$ connected by multiple nodes ${_w}\mathbf{n} \in \mathbf{R}^3$. The path $\tau$ satisfies the visibility condition $\mathcal{V}$ which is defined as:
\begin{equation}
\mathrm{state}({_w}\mathbf{m}) = \mathrm{Free}, 
\end{equation}
\begin{equation}
{_w}\mathbf{m} = {^*_w}\mathbf{n} + \alpha({_w}\mathbf{n} - {^*_w}\mathbf{n}), \ \alpha \in (0, 1),
\end{equation}
where $*$ indicates the parent node of ${_w}\mathbf{n}$. ${_w}\mathbf{m}$ is parameterized by $\alpha$, denoting an intermediate point on the line connected by node ${_w}\mathbf{n}$ and its parent node ${^*_w}\mathbf{n}$. The condition means that it is visible between each adjacent two nodes on the path $\tau$. 


For our obstacle-aware method, we propose the visual plane $\mathcal{H}_n$ for node ${_w}\mathbf{n}$.
Assuming the robot at the node ${_w}\mathbf{n}$ observing towards the target $_{w}\mathbf{p}_{end}$, its line of sight is occluded by a particular obstacle $\mathcal{C}_{\star}$ with a label $\star$. Accordingly, $\mathcal{H}_n$ is defined as:
\begin{equation}
\mathcal{H}_n = \{ {_w}\mathbf{h} \in \mathbf{R}^3 | ({_w}\mathbf{o}_{occ} - {_w}\mathbf{n})^{\mathrm{T}} ({_w}\mathbf{h} - {_w}\mathbf{o}_{occ}) = 0 \},
\end{equation}
where ${_w}\mathbf{h}$ is the point on the plane. $\mathcal{H}_n$ indicates the plane perpendicular to the view direction towards the target and passing through the occlusion position $_{w}\mathbf{o}_{occ}$ on the obstacle $\mathcal{C}_{\star}$. Then, we set up a polar coordinate in plane $\mathcal{H}_n$ with ${_w}\mathbf{o}_{occ}$ as the origin. Using transformation function $\rho:{_p}\mathbf{h} \rightarrow {_w}\mathbf{h}$, a 3D point ${_w}\mathbf{h}$ can be converted from a polar coordinate point ${_p}\mathbf{h} = (\theta, l)$ on plane $\mathcal{H}_i$ with the polar angle $\theta$ and radius $l$.

\begin{algorithm}[h]
\label{toposearch}
\SetAlgoNoEnd
\caption{Obstacle-aware Topological Path Search}
${_w}\mathbf{n}_{start}$ $\leftarrow$ {\bf Initialize}$()$; \\
$Q$.{\bf Enqueue}$({_w}\mathbf{n}_{start})$; \\
\While{$\neg Q$.{\bf empty}$()$}
{
  ${_w}\mathbf{n}_i$ $\leftarrow$ $Q$.{\bf Dequeue}$()$; \\
  $\iota$ $\leftarrow$ {\bf CheckVisibility}$({_w}\mathbf{n}_i, {_w}\mathbf{n}_{end})$; \\
  \If{$\iota == null$}
  {
    {\bf ExtendNode}$({_w}\mathbf{n}_i, {_w}\mathbf{n}_{end})$; \\
    {\bf continue}; \\
  }
  \Else
  {
    \If{$\iota ==$ Default}   
    {
      $\iota, \mathcal{C}_{\iota} \leftarrow$ {\bf RegionGrowth}($_{w}\mathbf{o}_{occ}$); \\
    }

    $\mathcal{H}_i$ $\leftarrow$ {\bf VisPlaneGenerate}$({_w}\mathbf{n}_i, \mathcal{C}_{\iota})$; \\
    
    \If{$\iota == $ {\bf CheckVisibility}$({^*_w}\mathbf{n}_i, {_w}\mathbf{n}_{end})$}
    {
      $(\theta_{k_{\mathbf{n}_i}})$ $\leftarrow$ {\bf SampleVisPlane}$(\mathcal{H}_i)$; \\
    }
    \Else
    {
      $(\theta_0,...,\theta_K)$ $\leftarrow$ {\bf SampleVisPlane}$(\mathcal{H}_i)$; \\
    }
    
    $({_w}\mathbf{n}_{i,0},...,{_w}\mathbf{n}_{i,J})$ $\leftarrow$ {\bf SearchNewNodes}$(\mathcal{H}_i)$; \\
    
    \For{${_w}\mathbf{n}_{i,j}$ in ${_w}\mathbf{n}_{i,0},...,{_w}\mathbf{n}_{i,J})$}
    {
      {\bf ExtendNode}$(n_i, n_{i,j})$; \\
      $Q$.{\bf Enqueue}$(n_{i,j})$; \\
    }
  }
}
$\{\tau_0,...,\tau_N\}$ $\leftarrow$ {\bf ConnectNodes}();
\end{algorithm}

\subsection{Obstacle-aware Topological Search}

As shown in Alg.~\ref{toposearch}, the obstacle-aware search is adopted for the topological graph construction. The algorithm starts by creating node $n_{start}$ at the start position $_{w}\mathbf{p}_{start}$ and adding it to the extension list $Q$ (Line 1-2). As the initial node, we have ${_w}\mathbf{n}_{start} = {_w}\mathbf{p}_{start}$. 

In each iteration, the topology graph expands from the nodes in $Q$. For a node ${_w}\mathbf{n}_i$ dequeued from the current extension list $Q$, we first check the visibility between ${_w}\mathbf{n}_i$ and the target $_{w}\mathbf{p}_{end}$ (Line 4-5). If occluded by point $_{w}\mathbf{o}_{occ}$ in the map and the index $\iota$ of the point is not defined, it means that an unknown object appears. In this case, we perform region growth at $_{w}\mathbf{o}_{occ}$ and obtain a point cloud cluster $\mathcal{C}_{\iota} \subset {_{w}M_{k}}$, where the region growth algorithm is realized by the fast NNS of IROP-Map. 
Each point in $\mathcal{C}_{\iota}$ is labeled with index $\iota$, which indicates the $\iota$-th encountered obstacle (Line 10-11). If $_{w}\mathbf{o}_{occ}$ has been labeled by $\iota$, it indicates an occlusion by an object that has been encountered before. Then, we search for the child nodes of ${_w}\mathbf{n}_i$ on the visual plane $\mathcal{H}_{i}$ (Line 12-17).

\begin{figure}[t]
\centerline{\includegraphics[scale=0.55]{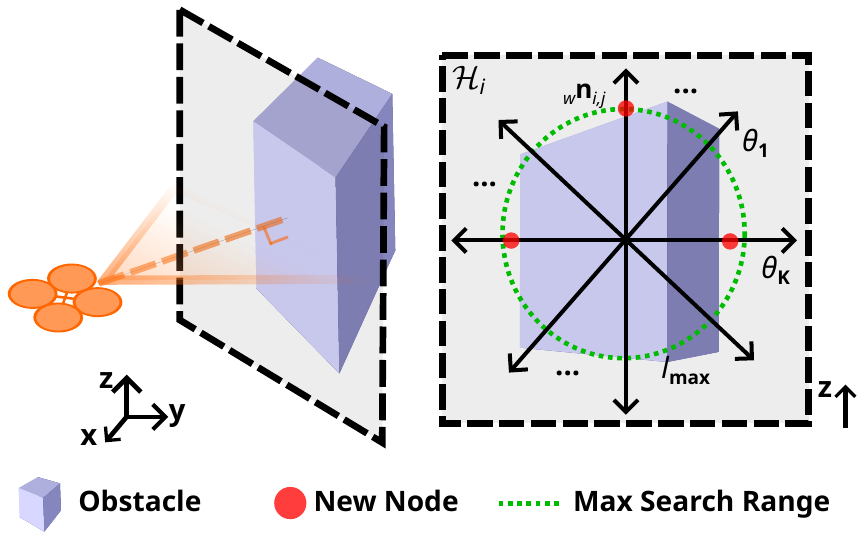}}
\caption{The illustration of the node extension process. While performing the node expansion process, a visual plane is defined first. The child nodes will be searched on the plane with a defined angle sample resolution.}
\label{f: h_plane}
\end{figure}

Subsequently, we sample on the visual plane $\mathcal{H}_{i}$ with $K$ angle values, as shown in Fig. \ref{f: h_plane}. For each distinct sampled angle $\theta_k = 2k\pi/K, k \in (1,...,K)$, we search for at most one possible existing child node ${_w}\hat{\mathbf{n}}_{i,k}$ that can be expanded. The process can be formulated as:
\begin{equation}
\min\limits_{l} \ \mathrm{state}(\rho({_p}\mathbf{h}_i^k)) = \mathrm{Free}, 
\end{equation}
\begin{equation}
\rho({_p}\mathbf{h}_i^k) = {_w}\hat{\mathbf{n}}_{i,k}, \ {_p}\mathbf{h}_i^k = (\theta_k, l), \ l < l_{max},
\end{equation}
where $l_{max}$ is the maximum search range. As a result, we can get $J$ expanded child nodes, where $J \leq K$. We record the sampled value of a node ${_w}\mathbf{n}$ while searching by $k_{\mathbf{n}}$. 
This method guarantees that each new node leads to a different topological space.
Specifically, if the obstacle occluding node ${_w}\mathbf{n}_i$ is the same as the obstacle occluding its parent node ${^*_w}\mathbf{n}_i$, the search of its child node will only proceed at the same distinct angle:
\begin{equation}
\theta_k = 2k\pi/K, \ k = k_{\mathbf{n}_i},
\end{equation}
while in this case $J \leq 1$ (Line 10-12). This is designed to avoid infinitely repeated searches near the same obstacle. A larger value of $K$ will result in more paths but also increase the computation time. We choose $K=4$ in this article to balance the performance. Hitherto, we connect all the new child nodes and add them to the extension list $Q$, and then end this search iteration (Lines 18-20).

\begin{figure}[t]
\centerline{\includegraphics[scale=0.715]{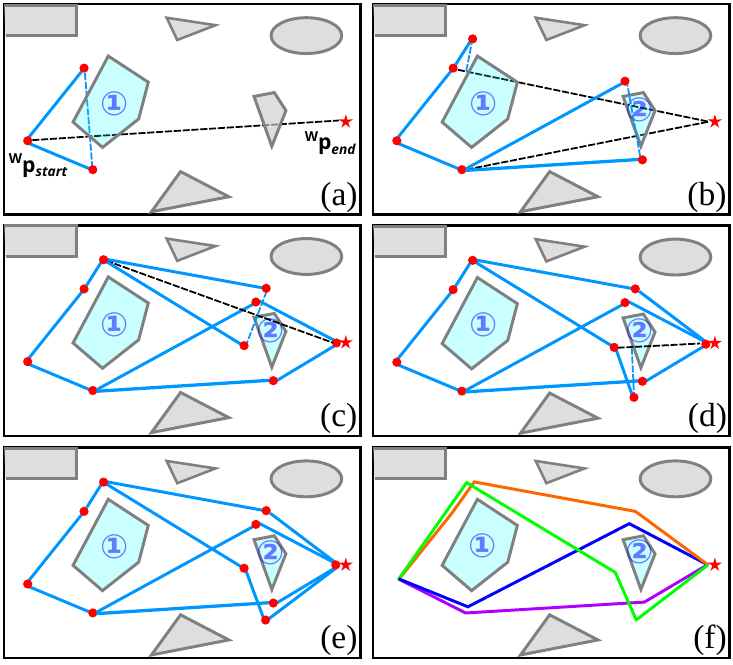}}
\caption{The illustration of the topological path search. (a) Encounter a new obstacle and expand the node in two directions. (b)(d) Encounter the same obstacle and expand the node in the previous direction. (c) Encounter a known obstacle and expand the node in two directions. (e) Complete the graph construction. (f) Capture the distinct paths.}
\label{f: topo_path}
\end{figure}

Finally, we back-propagate from the end nodes to connect the topology graph, and we can get $N$ distinct paths $\{\tau_0, ..., \tau_N\}$ (Line 21). The overall demonstration of our path planning method is shown in Fig. \ref{f: topo_path}.


\section{Adaptive Gradient-based High-speed Trajectory Planning}
In this section, we propose our adaptive gradient-based high-speed trajectory planning method. The adaptive trajectory initialization algorithm significantly improves the performance of trajectory generation, especially for the highly non-convex case of high-speed flights.
The optimization iterations are fewer and closer to the global optimal solution. For obstacle avoidance, we use our proposed RESDF in subsection~\ref{s:3D} for efficient and high-quality cost and gradient evaluation. Finally, we choose the best trajectory optimized from different topology references, which achieves a more approximate global optimum compared to existing works.

\subsection{Trajectory Parameterization}

For a differentially flat system such as the UAV, we first define its motion as a piece-wise 3-dimension and 5-degree polynomials $p(t)$ with $M$ pieces, and the $j$th piece can be expressed as: $p_{j}(t) = \mathbf{c}_{j}^{\mathrm T} \beta(t),  t \in [0,T_{j}]$, where $\mathbf{c}_{j} \in \mathbf{R}^{6 \times 3}$ is the coefficient matrix of the piece and $\beta(t) = [1, t, ..., t^{5}]^{\mathrm T}$ is natural basis vector. $T_{j}$ is the duration of the piece.

In this article, we adopt $\mathbf{MINCO}$ \cite{gcopter} class to achieve the spatial-temporal decoupled optimization. The trajectory $p(t)$ can only be parameterized by the time duration of each piece $\mathbf T = [T_{1}, ..., T_{M}]^{\mathrm T}$ and the intermediate waypoints $\mathbf q = [q_{1}, ..., q_{M-1}]^{\mathrm T}$ with a convenient space-time deformation $\mathcal M$: 
\begin{equation}
\mathbf c = \mathcal M(\mathbf q, \mathbf T),
\label{eq: minco}
\end{equation}
where $\mathbf c = [\mathbf c_{1}^{\mathrm T}, ..., \mathbf c_{M}^{\mathrm T}]$. 

\subsection{Adaptive High-speed Trajectory Initialization}

Although $\mathbf{MINCO}$ has the capability to optimize time directly, simple time pre-allocation methods (e.g., constant velocity or trapezoidal velocity method) will also lead to a bad convergence towards local minima in optimization. Especially for high-speed trajectories, the object function becomes highly non-convex. Unreasonable time allocation can lead to severe trajectory deformation and even the inability to find a solution.

\begin{figure}[h]
\centerline{\includegraphics[scale=1.0]{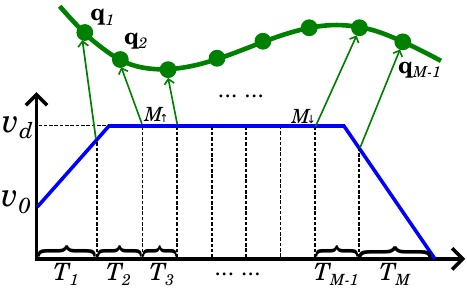}}
\caption{The illustration of the adaptive trajectory initialization. The allocation of the time vector is based on the desired speed and is approximate to temporal optimal.}
\label{f: time_init}
\end{figure}

Here, we propose our adaptive trajectory initialization method for high-speed $\mathbf{MINCO}$ trajectory, as shown in Alg.~\ref{traj_init}. In section IV, we obtain $N$ distinct topology paths.
Accordingly, each topology path will guide an initialization of a trajectory. 
For a $M$ piece trajectory, the initial waypoints $\mathbf q$ are sampled uniformly on its reference path $\tau$ with length $L$, and each piece with length $L/M$.
Then, we pre-allocate the time vector by a desired maximum flight speed $v_d$ and acceleration $a_d$. 
We assume that the UAV first accelerates from $v_0$ to $v_d$ with $a_d$ in $(1, M_{\uparrow})$ pieces and then flies with constant speed $v_d$. Finally, the UAV decelerates to 0 with acceleration $a_d$ in $(M_{\downarrow}, M)$ pieces. The process satisfies:
\begin{equation}
((v_d - v_0)^2 + v_d^2)/2a_d \leq L, 
\end{equation}
If $v_d$ does not meet this condition, we iteratively multiply $v_d$ by a scaling factor $0 < \gamma < 1$ until it satisfies the feasibility (Lines 1-2). Then, we can determine the three phases of the trajectory by $M_{\uparrow}$ and $M_{\downarrow}$ (Line 3):
\begin{equation}
M_{\uparrow} = \lceil M(v_d - v_0)^2/2La_d \rceil, \ M_{\downarrow} = \lceil M - Mv_d^2/2La_d \rceil, 
\end{equation}
Based on the boundary conditions, the duration of each piece of the trajectory can be solved accordingly, and we can obtain the overall initial time vector $\mathbf{T}$ (Lines 5-13). The illustration is shown in Fig. \ref{f: time_init} as well. 

\begin{algorithm}[h]
\label{traj_init}
\SetAlgoNoEnd
\caption{Adaptive Time Allocation}
\bf{Input}: $v_d, a_d, L, M$ \\
\bf{Output}: $\mathbf{T} = [T_{1}, ..., T_{M}]^{\mathrm T}$ \\
\While{\bf{CheckFeasibility}$(v_d)$ == false}
{
  $v_d$ $*=$ $\gamma$; \\
}
$M_{\uparrow}$, $M_{\downarrow}$ $\leftarrow$ \bf{GetHeadTailPieces}$(v_d, v_0)$; \\
$v_t = v_0$; \\
\For{$i$ in $(1,M_{\uparrow})$}
{
  $T_i$ $\leftarrow$ \bf{GetDuration}$(v_t, L/M)$; \\
  $v_t$ $\leftarrow$ \bf{UpdateVel}$()$; \\
}
$v_t = 0$; \\
\For{$i$ in $(M, M_{\downarrow})$}
{
  $T_i$ $\leftarrow$ \bf{GetDuration}$(v_t, L/M)$; \\
  $v_t$ $\leftarrow$ \bf{UpdateVel}$()$; \\
}
\For{$i$ in $(M_{\uparrow}, M_{\downarrow})$}
{
  $T_i$ $\leftarrow$ \bf{GetDuration}$(v_t, L/M)$; \\
}
\end{algorithm}

A better initial value of the trajectory brings this highly non-convex optimization closer to the optimal solution.
Hitherto, we can get multiple adaptive initial trajectories that pass through diverse collision-free regions, which leads to a more thorough coverage of the solution space and provides superior replanning compared with generating a single trajectory, especially while moving at a high speed.

\subsection{Gradient-Based Optimization}
For $\mathbf{MINCO}$, a general optimization problem can be formulated as:
\begin{equation}
\min\limits_{\mathbf{q},\mathbf{T}} \mathcal J = \mathcal L(\mathbf{q},\mathbf{T}) + \sum {\lambda_{\star}} \mathcal I_{\star}(\mathbf{c}(\mathbf{q},\mathbf{T}),\mathbf{T}),
\label{eq: J}
\end{equation}
\begin{equation}
\mathcal L = \int _{0}^{T_{t}} ||p^{(3)}(t)||^{2} \mathrm{d}t + \rho T_{t}, \quad T_{t} = \sum\limits_{j=1}^{M} T_{j},
\end{equation}
\begin{equation}
\mathcal I_{\star} = \sum\limits_{l=1}^{M} \frac{T_{j}}{\kappa_{l}} \sum\limits_{\tau = 0}^{\kappa_{j}} \max \{\mathcal{G}_{\star}(\mathbf{c}_{j},T_{j},\frac{\tau}{\kappa_{j}}), 0\}^3,
\end{equation}
where $\mathcal L$ is the time-regularized control effort, which minimizes the integral jerk and the total duration $T_{t}$ of the trajectory, and $\rho$ is the tunable weight value. $\mathcal I$ is the time integral penalty with weight $\lambda$, and $\kappa$ means the sample numbers on a piece of trajectory, $\frac{\tau}{\kappa}$ indicates the specific timestamp. $\mathcal{I}$ is determined by a specific cost function $\mathcal{G}$. $\mathcal{J}$ is the overall penalty.

To solve the optimization problem, we need the gradient of $\mathcal{J}$ w.r.t $\mathbf q$ and the gradient of $\mathcal{J}$ w.r.t $\mathbf T$, which can be derived by the Gradient Propagation Law:
\begin{equation}
\label{eq: pro}
\frac{\partial \mathcal{J}(\mathbf{q}, \mathbf{T})}{\partial \mathbf{q}}=\frac{\partial \mathcal{J}}{\partial \mathbf{c}} \frac{\partial \mathbf{c}}{\partial \mathbf{q}},\ 
\frac{\partial \mathcal{J}(\mathbf{q}, \mathbf{T})}{\partial \mathbf{T}}=\frac{\partial \mathcal{J}}{\partial \mathbf{T}}+\frac{\partial \mathcal{J}}{\partial \mathbf{c}} \frac{\partial \mathbf{c}}{\partial \mathbf{T}},
\end{equation}
where $\partial \mathbf{c} / \partial \mathbf{q}$ and $\partial \mathbf{c} / \partial \mathbf{T}$ can be derived from Eq.~\eqref{eq: minco}. Thus, this problem can be solved efficiently by unconstrained optimization algorithms such as L-BFGS.

Subsequently, we expect different types of cost function $\mathcal G_{\star}$ to indicate the requirements of the trajectory. The detailed penalty functions are designed as follows:

\emph{1) Obstacle avoidance}: In subsection \ref{s:3D}, we propose the concept of RESDF, where the distance to obstacle and gradient at a random position can be obtained directly by functions $\mathcal{D}$ and $\mathcal{F}$ in Eq.~\eqref{eq: dist} and Eq.~\eqref{eq: dist_grad}, respectively. Consequently, the obstacle collision cost $\mathcal{G}_{s}$ can be formulated as:
\begin{equation}
\mathcal{G}_{s} = d_{s} - \mathcal{D}(p_{j}(t_{\tau}),
\end{equation}
where $d_{s}$ is the safety threshold between the robot and the nearest obstacle. $t_{\tau} = T_{j} \frac{\tau}{\kappa_{j}}$ is a specific sample timestamp on the $j$th piece of the trajectory. Then, the gradient of $\mathcal{G}_{s}$ can be calculated by:
\begin{equation}
\frac{\partial \mathcal{G}_s}{\partial \mathbf{c}} 
= \frac{\partial \mathcal{G}_s}{\partial p} \frac{\partial p}{\partial \mathbf{c}} 
= -\beta(t)\mathcal{F}(p_{j}(t_{\tau}))^{\mathrm{T}},
\end{equation}
\begin{equation}
\frac{\partial \mathcal{G}_s}{\partial t} 
= \frac{\partial \mathcal{G}_s}{\partial p} \frac{\partial p}{\partial t} 
= -p_{j}^{(1)}(t_{\tau})^{\mathrm{T}} \mathcal{F}(p_{j}(t_{\tau})),
\end{equation}
Accordingly, it can be substituted to the final gradient propagation by Eq.~\eqref{eq: pro}.

\begin{figure}[h]
\centerline{\includegraphics[scale=1.0]{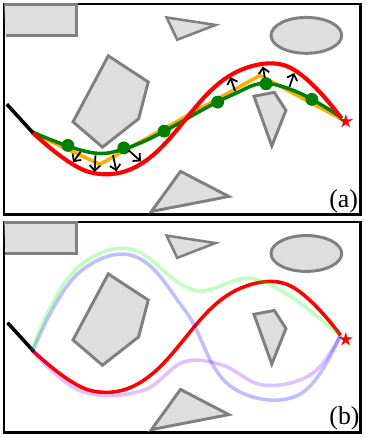}}
\caption{The illustration of the parallel gradient-based trajectory optimization. (a) The trajectory will be pushed away from the obstacles due to the gradient of RESDF. (b) Each distinct path leads to a trajectory generation, and the final trajectory will be selected with the smallest cost $\mathcal{J}$.}
\label{f: traj_opt}
\end{figure}

In most instances, the initial trajectory is collision-free due to the initialization along a safe path. Even though it is not completely collision-free, its major section has been attracted to free space. Hitherto, with the gradient descent based on RESDF, the trajectory will be pushed away from the obstacles and further increase safety, as shown in Fig. \ref{f: traj_opt}(a).


\emph{2) Dynamical feasibility}: The motion of the robot has to satisfy the dynamical feasibility. Here, we limit the range of velocity and acceleration by the cost function $\mathcal{G}_{v}$ and $\mathcal{G}_{a}$ expressed as follows:
\begin{equation}
\mathcal{G}_{v} = ||p_{j}^{(1)}(t_{\tau})||^{2} - v_{lim}^{2},
\end{equation}
\begin{equation}
\mathcal{G}_{a} = ||p_{j}^{(2)}(t_{\tau})||^{2} - a_{lim}^{2},
\end{equation}
where $v_{lim}$ and $a_{lim}$ are the maximum allowed magnitudes of velocity and acceleration. Similarly, the gradient of $\mathcal{G}_{v}$ and $\mathcal{G}_{a}$ can be derived by:
\begin{equation}
\frac{\partial \mathcal{G}_{v,a}}{\partial \mathbf{c}}
= \frac{\partial \mathcal{G}_{v,a}}{\partial p^{(n)}} \frac{\partial p^{(n)}}{\partial \mathbf{c}} 
= 2\beta^{(n)}(t)p_{j}^{(n)}(t_{\tau})^{\mathrm{T}},
\end{equation}
\begin{equation}
\frac{\partial \mathcal{G}_{v,a}}{\partial t} 
= \frac{\partial \mathcal{G}_{v,a}}{\partial p^{(n)}} \frac{\partial p^{(n)}}{\partial t} 
= 2p_{j}^{(n)}(t_{\tau})^{\mathrm{T}}p_{j}^{(n+1)}(t_{\tau}),
\end{equation}
where $n=\{1,2\}$, respectively. The dynamical feasibility will be guaranteed by using these two costs.

Above all, we can obtain multiple non-redundant trajectories in different local minima, while the trajectory with the smallest cost $\mathcal{J}$ 
will be chosen as the final trajectory, as shown in Fig. \ref{f: traj_opt} (b). Then, we solve the control commands from the trajectory that can be executed by the controller.

\begin{figure}[h]
\centerline{\includegraphics[scale=0.62]{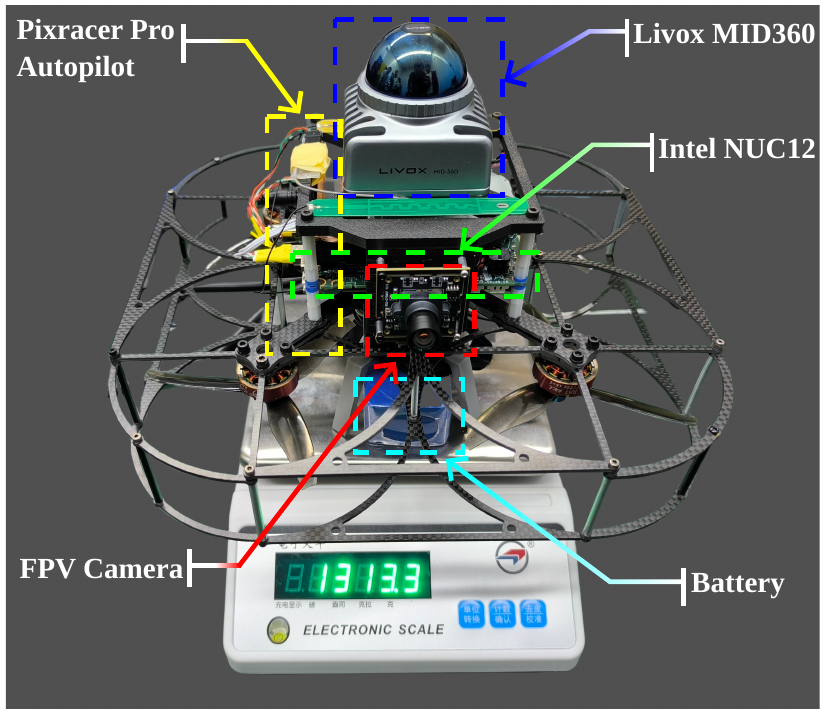}}
\caption{Hardware platform of our system.}
\label{f: uav}
\end{figure}

\section{Evaluations and Experiments}

\subsection{Implementation Details}

In this section, we present our hardware platform for experiments and evaluations. We design a 200mm wheelbase quadrotor frame with the protection of the rotors, carrying an Intel NUC12WSH-i7 running Ubuntu 20.04 as the onboard computer, and a Livox MID-360 Lidar with an FOV of 360°(horizontal)×59°(vertical) and detection range of 40m
within 10\% reflectivity is equipped for onboard sensing, while publishing point cloud at 10Hz. The UAV controller is a Micoair-NXT board that runs the PX4 flight stack. 
The overall system weighs 1.31 kg.  The overview of our hardware platform is shown in Fig. \ref{f: uav}. The following simulations and evaluations of the algorithms are also conducted on Intel NUC12WSHi7.

\subsection{Evaluations of Incremental Local Mapping}

\begin{figure*}[t]
\centerline{\includegraphics[scale=1.01]{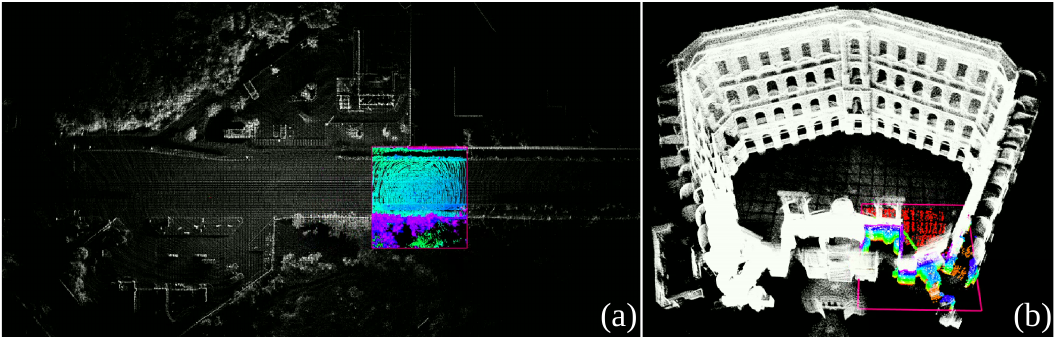}}
\caption{The presentation of our mapping method in indoor and outdoor scenarios. (a) is the outdoor Kitti\_s04 dataset recorded by VLP-16 on an urban road. (b) is the indoor dataset recorded by MID-360 in an HK building corridor.}
\label{f: map_demo}
\end{figure*}

\begin{table*}[t]
\centering
\caption{
Benchmark Comparison of Local Mapping}
\begin{tabular}{cccccccccc}
   \toprule
   \multicolumn{2}{c}{\textbf{Resolution}(m)} & \multicolumn{4}{c}{0.1} & \multicolumn{4}{c}{0.2} \\
  \hline
  \textbf{Dataset} & \textbf{Method} &  $t_{map}$(ms)&$t_{update}$(ms)& $t_{esdf}$(ms)& $m$(MB)&  $t_{map}$(ms)&$t_{update}$(ms)& $t_{esdf}$(ms)& $m$(MB)   \\
   \hline
   \multirow{4}{*}{HK}& Ours  &  \textbf{3.78}&\textbf{3.78}& \textbf{0}& \textbf{39.68}&  \textbf{1.88}&1.88& \textbf{0}& \textbf{35.68}\\
 \multirow{4}{*}{Corridor}& ROG-Map\cite{rogmap} & 37.28&4.32& 32.96& 121.80& 6.77&\textbf{1.28}& 5.49& 77.03\\     
 & FIESTA\cite{fiesta} & 98.76&10.63& 88.13& 4342.25  &  26.93&4.28& 22.65&603.08\\     
 & OctoMap\cite{octomap} & 327.09&197.17& 135.92& 476.15  &  84.62&36.52& 48.10&294.33\\
 \hline
   \multirow{4}{*}{Kitti\_04}& Ours                      & \textbf{18.34}&18.34& \textbf{0}& \textbf{78.94}  &  \textbf{6.73}&\textbf{6.73}& \textbf{0}&\textbf{56.77}\\
 & ROG-Map\cite{rogmap} & 174.97&\textbf{13.52}& 161.45& 442.58  &  25.79&8.00& 17.79&248.12\\     
 & FIESTA\cite{fiesta} & 389.59&28.80& 360.79& 12901.67  &  59.71&18.98& 40.73&1678.84\\     
 & OctoMap\cite{octomap} & 1389.19&890.38& 498.81& 2092.85  &  309.72&185.26& 124.46&1306.51\\
 \bottomrule
\end{tabular}
\label{t: mapping_benchmark}
\end{table*}

To verify the advantage of our proposed IROP-Map with RESDF, we compare our algorithm with other three SOTA mapping methods: ROG-Map\cite{rogmap}, which is a robocentric occupancy grid map, FIESTA\cite{fiesta}, which is with a fixed map origin, and OctoMap\cite{octomap}, which is an octree-based occupancy grid map. All the three methods can generate ESDF based on their map structure. 

The evaluations are conducted on two datasets, as shown in Fig. \ref{f: map_demo}: \emph{1)} HK Corridor, an indoor dataset recorded by livox mid-360 lidar, with a size of 45 m * 45 m * 30 m. The local map range is set as 15 m * 15 m * 6 m in this scenario. \emph{2)} Kitti\_04, an outdoor dataset recorded by VLP-16 lidar on an urban road, with a size of 200 m * 100 m * 30 m. The local map range is set as 30 m * 30 m * 6 m in this scenario. Then, we test the methods on the datasets at 0.1 m and 0.2 m map resolution, respectively. We summarize the average time cost of map update $t_{update}$, the average time cost of ESDF update $t_{esdf}$, the average total time cost of mapping process $t_{map}$, and the average memory consumption $m$.

The comparison results are presented in Table \ref{t: mapping_benchmark}, with the best results highlighted in bold fonts. 
Our method can update the map with a very low time cost, even with the increase of the local map size and resolution. Although in some cases, ROG-Map has less time cost due to the efficiency of direct grid update, it still requires additional computation to obtain the ESDF. Our method totally eliminates the generation of ESDF, while we can still calculate the distance and gradient value at any position directly based on our unique map structure. With the same functionality output, our computational efficiency is significantly improved (only taking 10. 5\% of the time spent by the second-best method). The mapping time is approximately tenth of the fastest one in existing methods.

From the perspective of memory consumption, ROG-Map allocates the memory for each grid in the robocentric map, FIESTA allocates the memory for each grid within a fixed map size, and OctoMap maintains the nodes in a fixed map size by an octree. Our method only maintains the point cloud in the robocentric local map by $i$-$Octree$. This enables our method to run in large-scale scenarios with very low memory consumption. In general, our method is the most efficient in time and memory compared to the other SOTA methods. Such a low computational load is a great advantage for high-speed and large-scale flights, and the RESDF is easy to use for obstacle avoidance. In the deployment of the entire navigation system, we use a local map of 15 m * 15 m size with 0.1 m resolution to balance perception range and time efficiency.

\subsection{Evaluations of Obstacle-aware Path Search}
In this section, we validate the advantage of our proposed obstacle-aware topological path search algorithm. Our method is compared with three SOTA topological planning works: Raptor \cite{raptor}, Ego-PlannerV2 \cite{egov2}, and OAPR \cite{oapr}. The simulated environment is a 50 m * 50 m * 8 m size space with 150 column obstacles and 100 circle obstacles. The start position is the origin point of the map (0.0 m, 0.0 m, 0.0 m), and the target is a random point at a distance of 7.5 m or 15 m.

\begin{table}[h]
\centering
\caption{
Benchmark Comparison of Topological Path Search}
\begin{tabular}{ccccc}
   \toprule
      \textbf{Distance}(m) & \multicolumn{2}{c}{15} & \multicolumn{2}{c}{7.5}\\ 
   \hline
   \textbf{Method} &$t_{path}$(ms)& $N_{p}$& $t_{path}$(ms)&$N_p$\\
   \hline
    Ours                      & \textbf{1.57}&\textbf{7.75}& 0.49&\textbf{4.80}\\
  Raptor\cite{raptor}& 8.89&6.25& 6.11&4.25\\     
  Ego-PlannerV2\cite{egov2}& 2.17&1.60 & \textbf{0.35}&1.20\\     
  OAPR\cite{oapr}& 16.37&3.45 & 11.92&3.35\\
  \bottomrule
\end{tabular}
\label{t: topo_benchmark}
\end{table}

\begin{figure}[h]
\centerline{\includegraphics[scale=0.53]{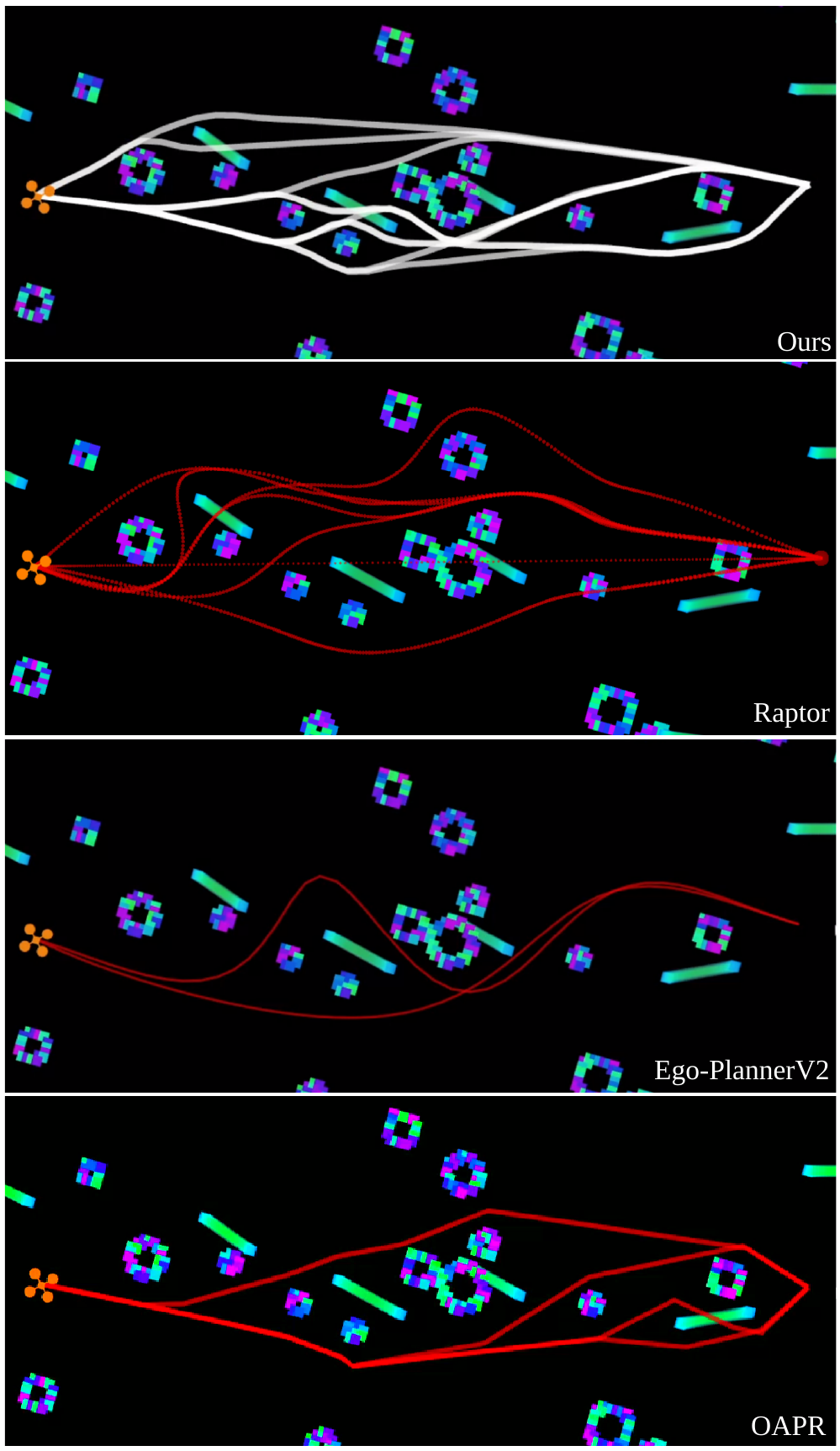}}
\caption{Comparison of the topological path search. Our proposed method can find the most candidate paths and has the shortest time cost.}
\label{f: topo_comp}
\end{figure}

The results are summarized by 20 tests and can be found in Table \ref{t: topo_benchmark}. $t_{path}$ is the average time cost of path planning, and $N_p$ is the average number of output candidate paths. A larger value of $N_p$ indicates higher space coverage and a higher probability of finding the global optimal. Raptor plans the paths by sampling in the space and PRM graph search, and the use of path shortcuts and pruning to get distinct paths, which are very time-consuming. OAPR can generate distinct paths directly, however, the process of clustering brings a high computational load, and this method makes it hard to tackle the environment with non-convex obstacle geometries, such as circles. Ego-PlannerV2 uses multiple A* modules to get distinct paths, which is less time-consuming compared to the above two methods, while only a very small number of candidate paths can be found. Our algorithm can find the most candidate paths in all the tests, meanwhile, takes the shortest time in most cases. The fast and high-quality front-end path planning leads the trajectory generation to be optimal in a larger space, which is significant for highly non-convex optimization while the trajectory speed is high. A demonstration of the comparison of these methods is shown in Fig. \ref{f: topo_comp}. 

\subsection{Evaluations of High-speed Flights in Simulation}

\begin{table*}[h]
\centering
\caption{
Benchmark Comparison of High-speed Planning System}
\begin{tabular}{cccccccccc}
   \toprule
   \textbf{Method}  &$v_{lim}$(m/s)&  $t_{total}$(ms)& $t_{map}$(ms)& $t_{path}$(ms)&$t_{traj}$(ms) &$v_{max}$(m/s)  &$l_{traj}$(m) &$T_{traj}$(s)&$\eta$(\%)\\
   \hline
    Ours                       &\multirow{5}{*}{5}&  \textbf{4.53}&\textbf{3.53}& 0.42&\textbf{0.78}&  \textbf{4.74}& \textbf{54.34}&\textbf{15.34}&\textbf{100}\\
  Raptor\cite{raptor}  &&  26.98&14.36& 6.65&5.97&  4.51& 55.06&16.38&100\\     
  Ego-PlannerV2\cite{egov2}  &&  8.35&5.35& \textbf{0.37}&2.63&  4.71& 54.44&17.08&95\\
  Fast-Dodge\cite{fastdodge}& & 11.68& 5.92& 1.04& 5.72& 4.15& 55.17& 17.23&95\\
  Agile-Auto\cite{learningsr}& &  25.87& /& /& /& \textcolor{red}{5.45}& 80.31& 13.56&75\\
 \hline
     Ours                       &\multirow{3}{*}{10}&  \textbf{5.44}&\textbf{3.44}& \textbf{0.45}&\textbf{1.55}&  \textbf{9.28}& \textbf{54.30}&\textbf{8.51}&\textbf{100}\\
  Raptor\cite{raptor}  &&  28.97&14.53& 8.08&6.36&  8.85& 54.68&10.22&60\\
  Agile-Auto\cite{learningsr}& & 26.52& /& /& /& \textcolor{red}{10.68}& 85.64& 8.64&40\\
  \hline
     Ours                       &\multirow{2}{*}{15}&  \textbf{5.46}&\textbf{3.24}& \textbf{0.42}&\textbf{1.80}&  \textbf{13.95}& \textbf{54.11}&\textbf{5.31}&\textbf{90}\\
 Agile-Auto\cite{learningsr}& & 26.30& /& /& /& 11.23& 82.69& 7.30&35\\
 \bottomrule
\end{tabular}
\label{t: planning_benchmark1}
\end{table*}

\begin{table*}[h]
\centering
\caption{
More challenging tests of our planning system in simulation}
\begin{tabular}{cccccccccc}
   \toprule
   \textbf{Scenario}  &$v_{lim}$(m/s)&  $t_{total}$(ms)& $t_{map}$(ms)& $t_{path}$(ms)&$t_{traj}$(ms) &$v_{max}$(m/s)  &$l_{traj}$(m) &$T_{traj}$(s)&$\eta$(\%)\\
   \hline
    1&15&  7.22&4.49& 0.86&1.87&  13.82& 55.17&5.99&85\\
 \hline
     2&15&  6.98&4.38& 0.53&2.07&  14.07& 145.34&17.65&80\\
  \hline
     3&5&  5.95&4.21& 0.51&1.23&  4.65& 54.42&15.69&90\\
  \bottomrule
\end{tabular}
\label{t: planning_benchmark2}
\end{table*}

\begin{figure}[h]
\centerline{\includegraphics[scale=0.535]{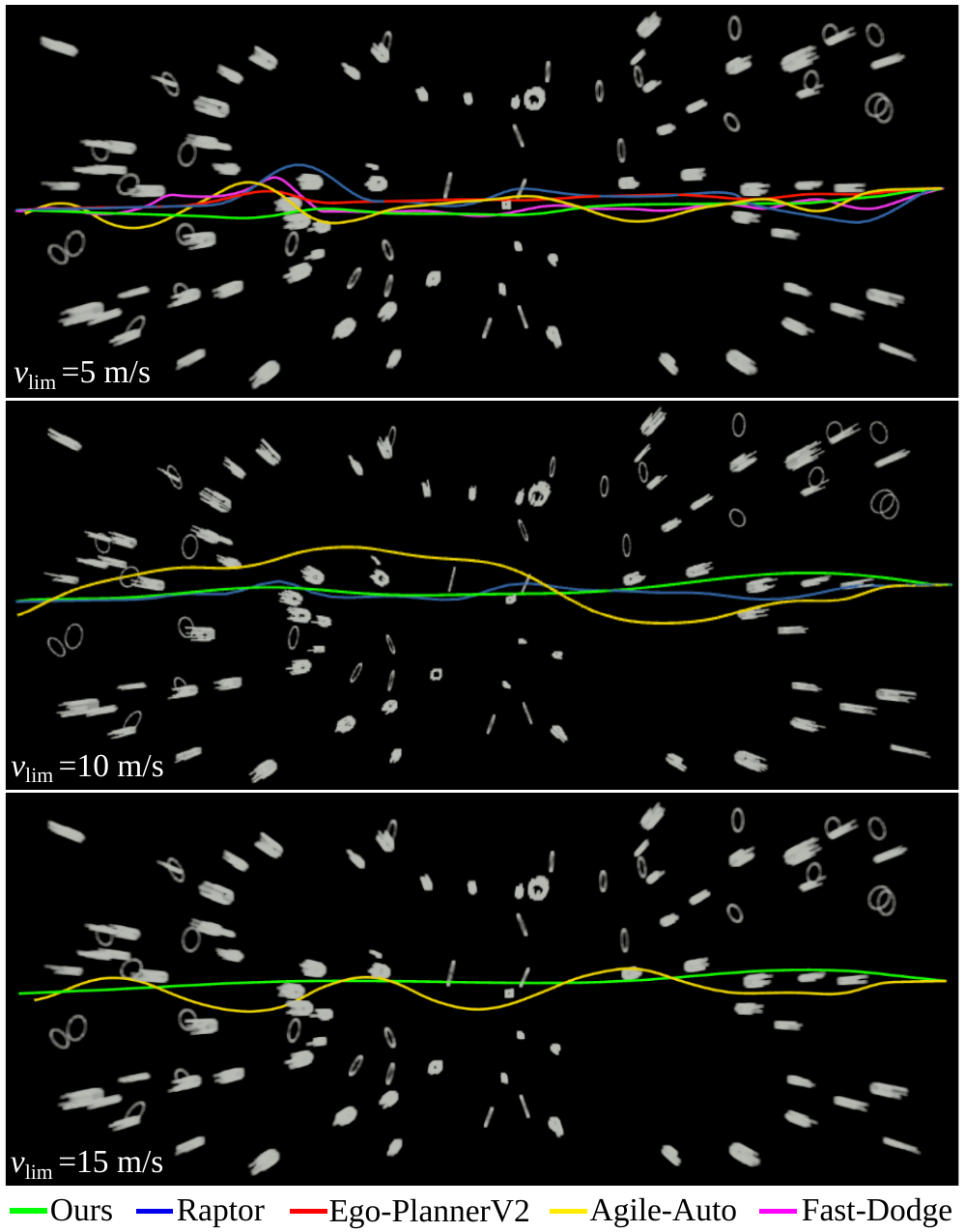}}
\caption{Trajectory benchmark comparison with the speed limitation as 5 m/s, 10 m/s, and 15 m/s.}
\label{f: traj_comp}
\end{figure}

In this section, we present the results of the entire high-speed navigation system in simulation. Firstly, we choose four SOTA works as baselines for comparison. Raptor \cite{raptor}, Ego-PlannerV2 \cite{egov2}, Fast-Dodge \cite{fastdodge} and Agile-Auto \cite{learningsr}. The evaluations are in a 50 m * 20 m * 8 m simulated forest with 80 column obstacles and 50 circle obstacles. We let the UAV fly from (-27.0 m, 0.0 m, 1.0 m) to (27.0 m, 0.0 m, 1.0 m) to evaluate the performance, and we divide the tests into three groups with speed limits $v_{lim}$ = 5 m/s, 10 m/s and 15 m/s, with each group run 20 times. We summarize the average mapping time cost $t_{map}$, the average front-end path search time cost $t_{path}$, the average trajectory optimization time cost $t_{traj}$, the average total time cost $t_{total}$, the average maximum speed $v_{max}$, the average trajectory length $l_{traj}$, the average trajectory duration $T_{traj}$ and the success rate $\eta$. The results are shown in Table \ref{t: planning_benchmark1} and Fig. \ref{f: traj_comp}. 

Raptor adopts an occupancy grid map with ESDF, PRM-based topological path search, and uses B-spline to formulate the trajectory. This method is the most time-consuming due to the ESDF and PRM modules. Moreover, the trajectory is spatial-temporal coupled, making it difficult to optimize trajectories successfully at high speeds. Ego-PlannerV2 is ESDF-free, and the A*-based topological path search is also more time-efficient. However, inadequate front-end paths make the trajectory hard to approach global optimality. Especially when the speed is high, the $\mathbf{MINCO}$ trajectory becomes highly non-convex. Without a suitable initial value, the optimizer can hardly find a solution. Fast-Dodge uses a kinodynamic path search, which makes it difficult to find high-quality paths at high speeds due to limitations in the sampling resolution limitations in the control space, and the back-end trajectory optimization faces the same challenges as Ego-PlannerV2. Agile-Auto is an end-to-end method based on imitation learning. The end-to-end policy can run in real time, but due to limitations in the training environment, this method lacks sufficient generalization and robustness and lacks the flexibility to adjust the expected speed arbitrarily. The red font indicates the case of error. In the case of $v_{lim}$ = 5 m/s, all the methods can work relatively robustly. As $v_{lim}$ increases to 10 m/s, Raptor and Agile-Auto can run at a low success rate. Both Ego-PlannerV2 and Fast-Dodge fail in this group of tests. Although Agile-Auto has the shortest time in some tests, however, the method cannot assure the speed limitation and exceed the desired speed.
When the speed limitation is 15 m/s, our system can still run robustly in all three groups and has a nearly 100$\%$ success rate. Agile-Auto performs a low success rate and cannot reach the desire speed due to the training limitation. Meanwile, all the other baselines fail in this group. 
Meanwhile, our system also achieves the lowest algorithm latency, the highest flight speed, the lowest flight duration, and the shortest flight trajectory length.

Then, to verify the improvement of our adaptive trajectory initialization method, we compare it with two of the most used methods: 1) constant velocity time allocation. The trajectory is hypothesized as a constant-velocity motion, and the duration of each piece is calculated by the length of the trajectory piece and the desired speed. 2) trapezoidal velocity time allocation. The method assumes that in every piece, the trajectory accelerates to the desired speed and then decelerates to zero. The comparison is also performed in the same simulation environment as above, and the speed limit is set to 15 m/s. As can be seen in Table \ref{t: traj_abla}, the initialization using constant velocities causes the initial values of majority trajectories violating constraints. Thus, the optimization fails many times, and even if the solution is successful, the trajectory might not be feasible. The trapezoidal velocity time allocation method is overly conservative. The speed of the output trajectory cannot reach the desired speed, which is far away from optimal. Our proposed method guarantees this highly non-convex optimization closer to the optimal solution, which significantly enhanced the performance of high-speed trajectories.

\begin{table}[h]
\centering
\caption{
Comparison of different initialization methods}
\begin{tabular}{ccccc} 
   \hline
   \textbf{Method}  &$v_{lim}$(m/s) &$T_{traj}$(s)& $v_{max}$(m/s)& $\eta$(\%)\\
   \hline
    Ours    &\multirow{3}{*}{15}& 5.31&\textbf{13.95}& \textbf{90}\\
    constant && \textbf{5.11}&\textcolor{red}{17.83}& 20\\     
    trapezoidal && 8.93&8.77& 90\\
  \bottomrule
\end{tabular}
\label{t: traj_abla}
\end{table}

\begin{figure}[h]
\centerline{\includegraphics[scale=0.425]{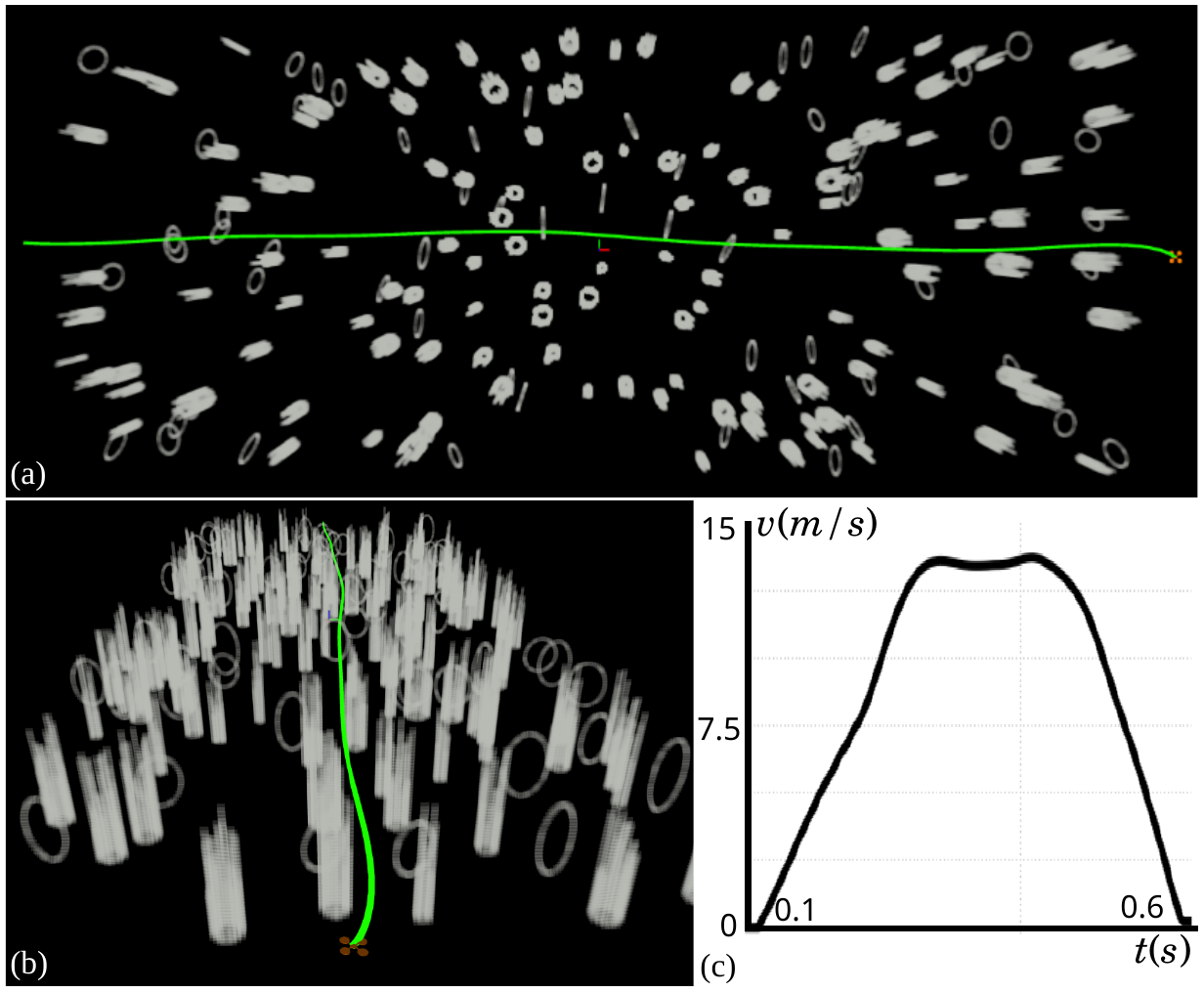}}
\caption{More challenging high-speed navigation in an extremely dense environment. When the speed limitation is set as 15 m/s, only our planning system can run in this environment. The maximum speed of the UAV reaches 13.8 m/s.}
\label{f: sim1}
\end{figure}

\begin{figure}[h]
\centerline{\includegraphics[scale=0.425]{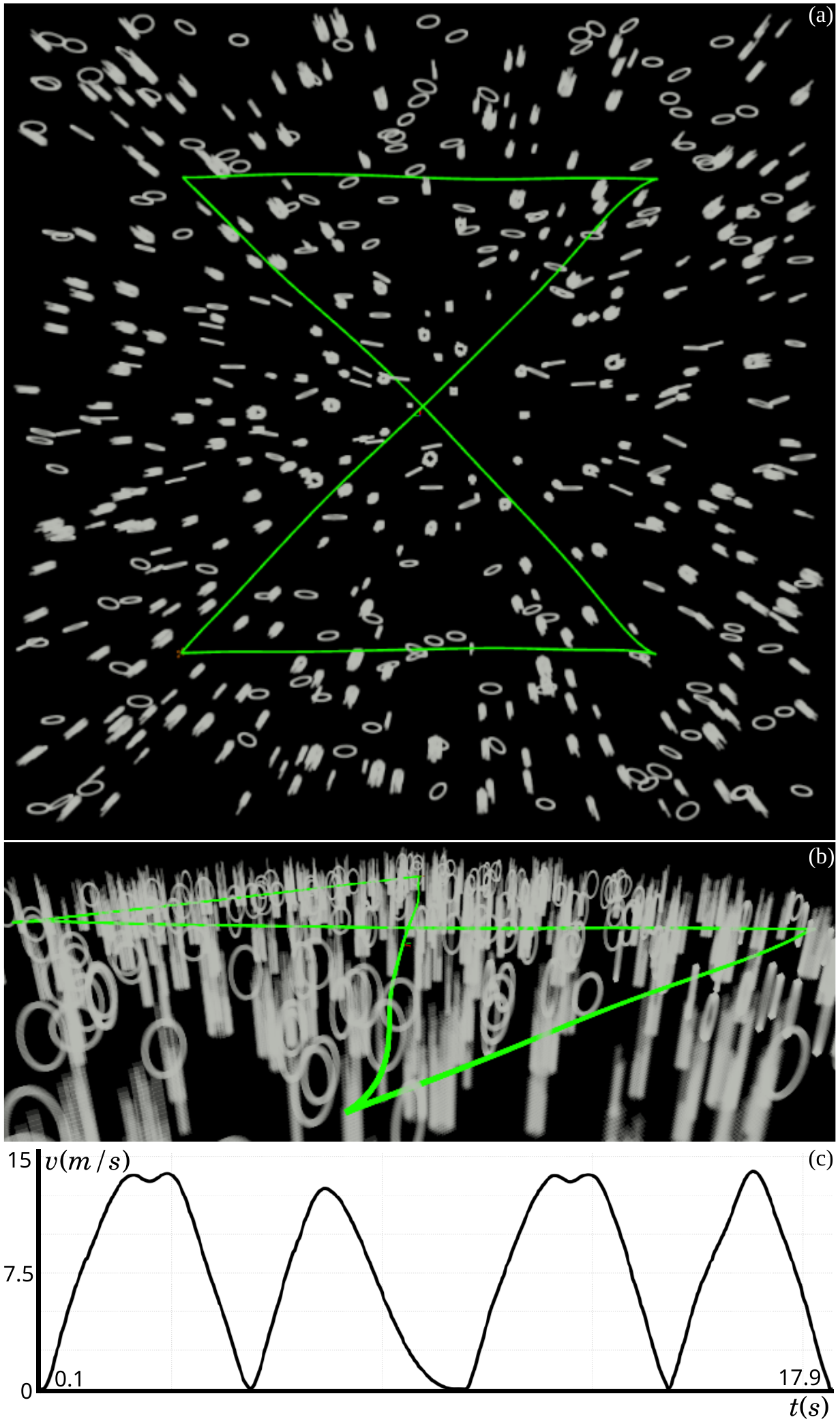}}
\caption{More challenging high-speed patrol in an extremely dense environment. When the speed limitation is set as 15 m/s, only our planning system can run in this environment. The maximum speed of the UAV reaches 14.1 m/s.}
\label{f: sim2}
\end{figure}

\begin{figure}[h]
\centerline{\includegraphics[scale=0.425]{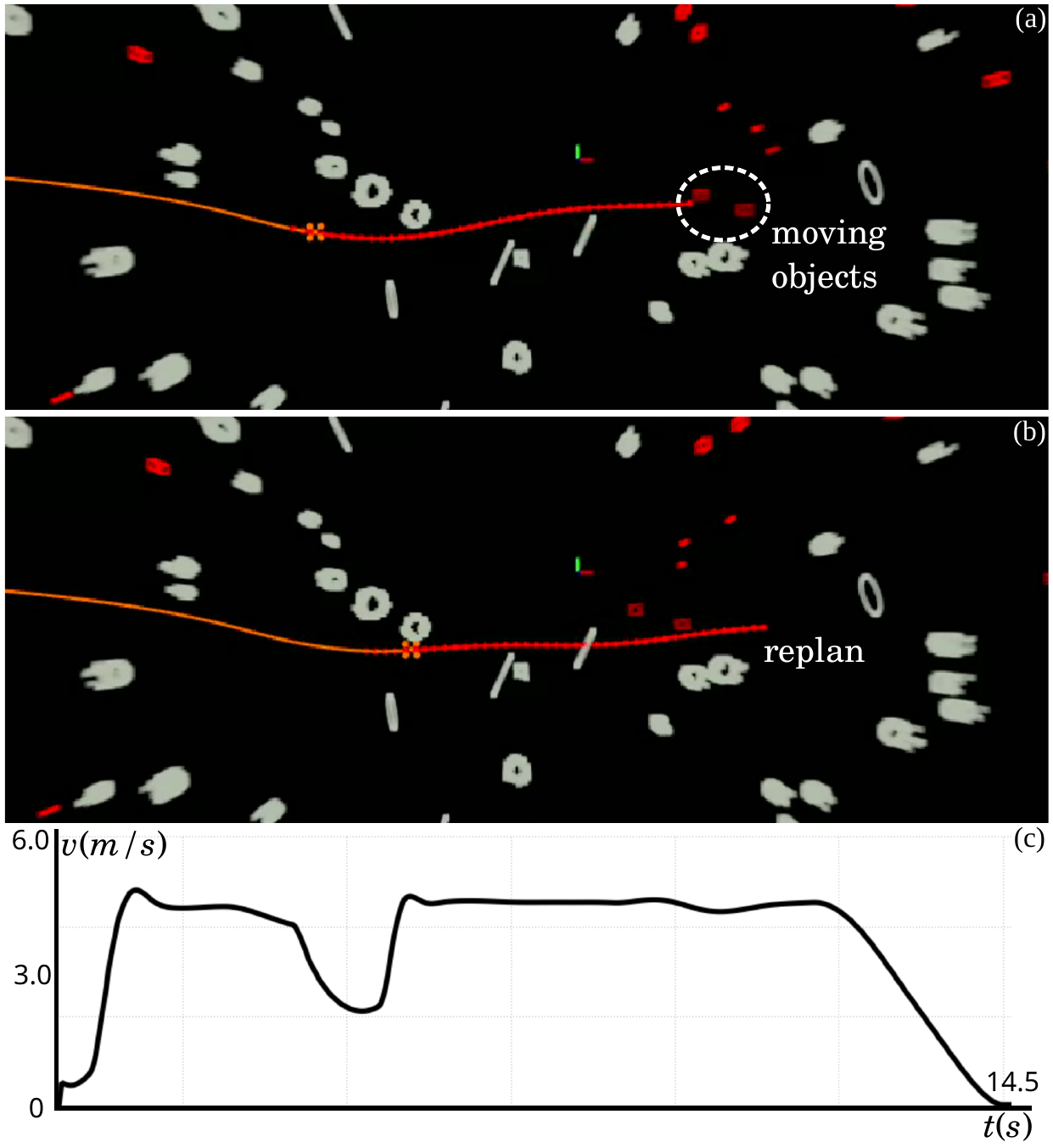}}
\caption{More challenging flight in a dynamic environment with red color denoting dynamic obstacles. The UAV can replan very fast and can tackle the dynamic obstacles.}
\label{f: sim3}
\end{figure}

Moreover, we demonstrate the advantage of our system in three more challenging scenarios, and the results are shown in Table \ref{t: planning_benchmark2}. Only our system can successfully run in these types of settings: \emph{1)} High-speed navigation in a very dense forest. This scenario is 50 m * 20 m * 8 m, with 150 columns and 100 circles, as shown in Fig. \ref{f: sim1}. Compared to the benchmark comparison environment, this scenario has three times the density of obstacles. The position of obstacles is random in each test. The UAV should navigate from (-27.0 m, 0.0 m, 1.0 m) to (27.0 m, 0.0 m, 1.0 m) with the speed limit $v_{lim} =$ 15 m/s. Our algorithm succeeded 17 out of 20 tests, with an average trajectory length of 55.17 m, a flight duration of 5.99 s, a latency in the algorithm of 7.22 ms, and the average maximum speed reaches 13.82 m/s. \emph{2)} High-speed patrol in a very dense forest. This scenario is 50 m * 50 m * 8 m with 300 columns and 200 circles, which also has nearly three times as many obstacles as before. The result is shown in Fig. \ref{f: sim2}. The position of obstacles is random in each test. The UAV starts at (-15.0 m, 15.0 m, 1.0 m) and is required to fly through 4 waypoints (15.0 m, -15.0 m, 1.0 m), (15.0 m, 15.0 m, 1.0 m), (-15.0 m, -15.0 m, 1.0 m) and (-15.0 m, 15.0 m, 1.0 m) in sequence. The speed limit is also set as 15 m/s. Among 20 tests, our system achieves a success rate of 80$\%$ with an average maximum speed of 14.07 m/s, 145.34 m average trajectory length, 17.65 s flight duration, and 6.98 ms algorithm latency. \emph{3)} Navigation in a corridor with cluttered fast-moving objects. This scenario is a 50 m * 20 m * 8 m corridor with 150 fast-moving objects. The speed of the objects varies randomly from 0.5 m/s to 2.0 m/s. The UAV is expected to navigate from (-27.0 m, 0.0 m, 1.0 m) to (27.0 m, 0.0 m, 1.0 m) with the speed limit $v_{lim}$ = 5 m/s. In this case, our system can still maintain a 90$\%$ success rate in 20 tests. The fast and robust planning allows the UAV to react instantly in the highly dynamic environment and achieves a 54.42 m average trajectory length, 15.69 s flight duration, 5.95 ms algorithm latency, and the average maximum speed a 55.17 m average trajectory length, 5.99 s flight duration, 7.22 ms algorithm latency and 4.65 m/s average maximum speed. Fig. \ref{f: sim3} shows an example of dynamic obstacle avoidance.

Based on all the quantitative studies in simulation, we can conclude that the whole system has the ability to run in real-time and guarantee safety in high-speed flights.

\subsection{High-speed Flights in Real-world Cluttered Environments}

Based on the simulation and quantitative studies, we finally validate our whole system in real-world experiments. Among all tests, the UAV uses the lidar inertial odometry \cite{fastlio} for state estimation. 

\begin{figure}[h]
\centerline{\includegraphics[scale=0.43]{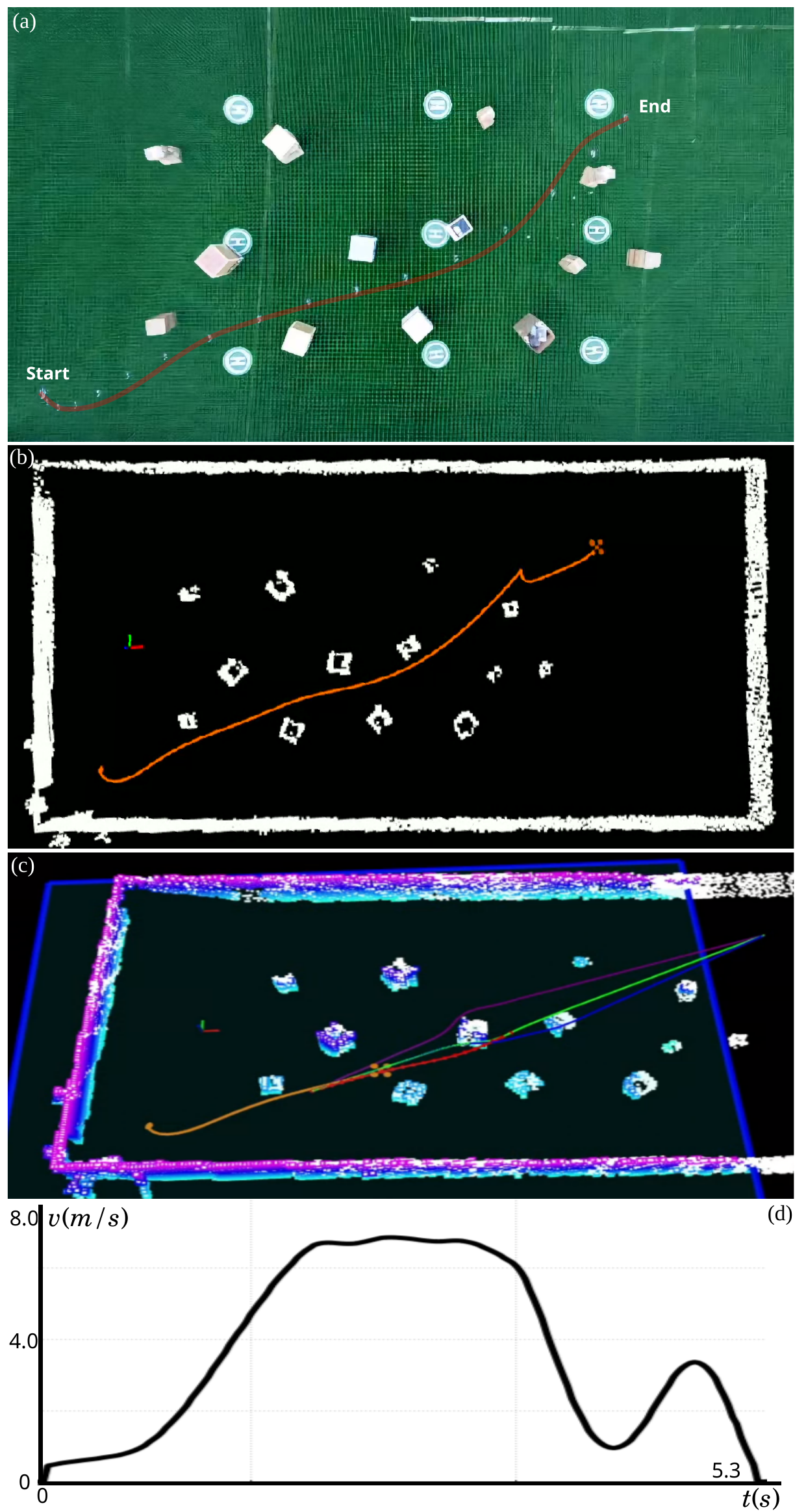}}
\caption{The results of an indoor high-speed navigation flight. (a) is the composed image of the flight path. (b) is the rviz visualization of the flight history. (c) is the rviz visualization of the planning process. (d) is the speed curve of the UAV.}
\label{f: realworld4}
\end{figure}

\begin{figure}[h]
\centerline{\includegraphics[scale=0.43]{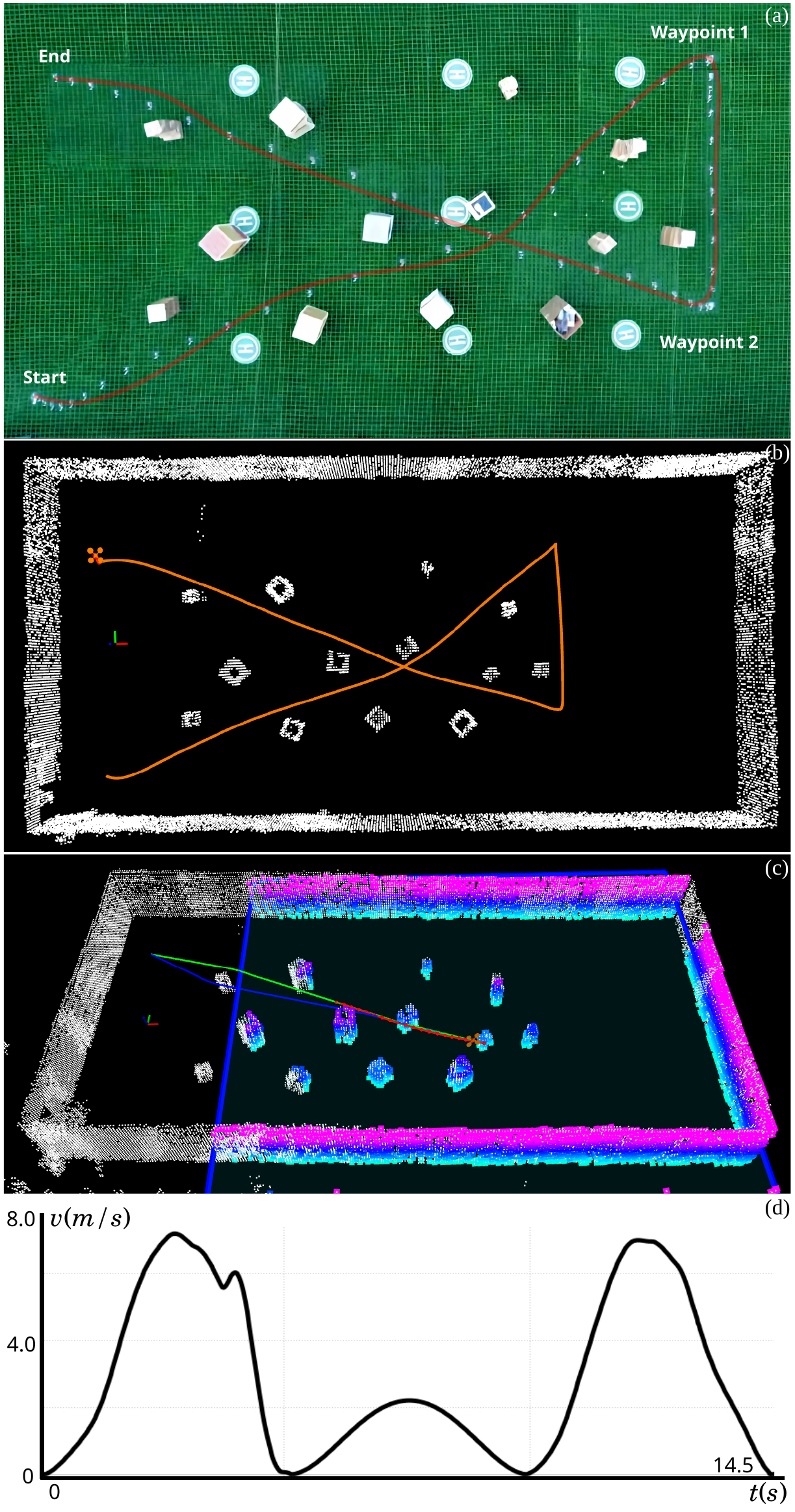}}
\caption{The results of an indoor high-speed patrol flight. (a) is the composed image of the flight path. (b) is the rviz visualization of the flight history. (c) is the rviz visualization of the planning process. (d) is the speed curve of the UAV.}
\label{f: realworld1}
\end{figure}

We first test our system in an indoor cluttered room with 14 obstacles in it. The speed limit $v_{lim}$ is set as 8.5 m/s as the obstacle density is high in this scenario.
Firstly, we let the UAV navigate from the start point to the goal, as shown in Fig. \ref{f: realworld4}. Then, the UAV is required to patrol in the room and pass three waypoints, as shown in Fig. \ref{f: realworld1}. In several repetitive experiments, our UAV is able to complete the task successfully every time. The system outputs an optimized trajectory based on fast perception and multi-path guided trajectory generation, and the maximum speed of our UAV is over 8.0 m/s.

\begin{figure}[h]
\centerline{\includegraphics[scale=0.43]{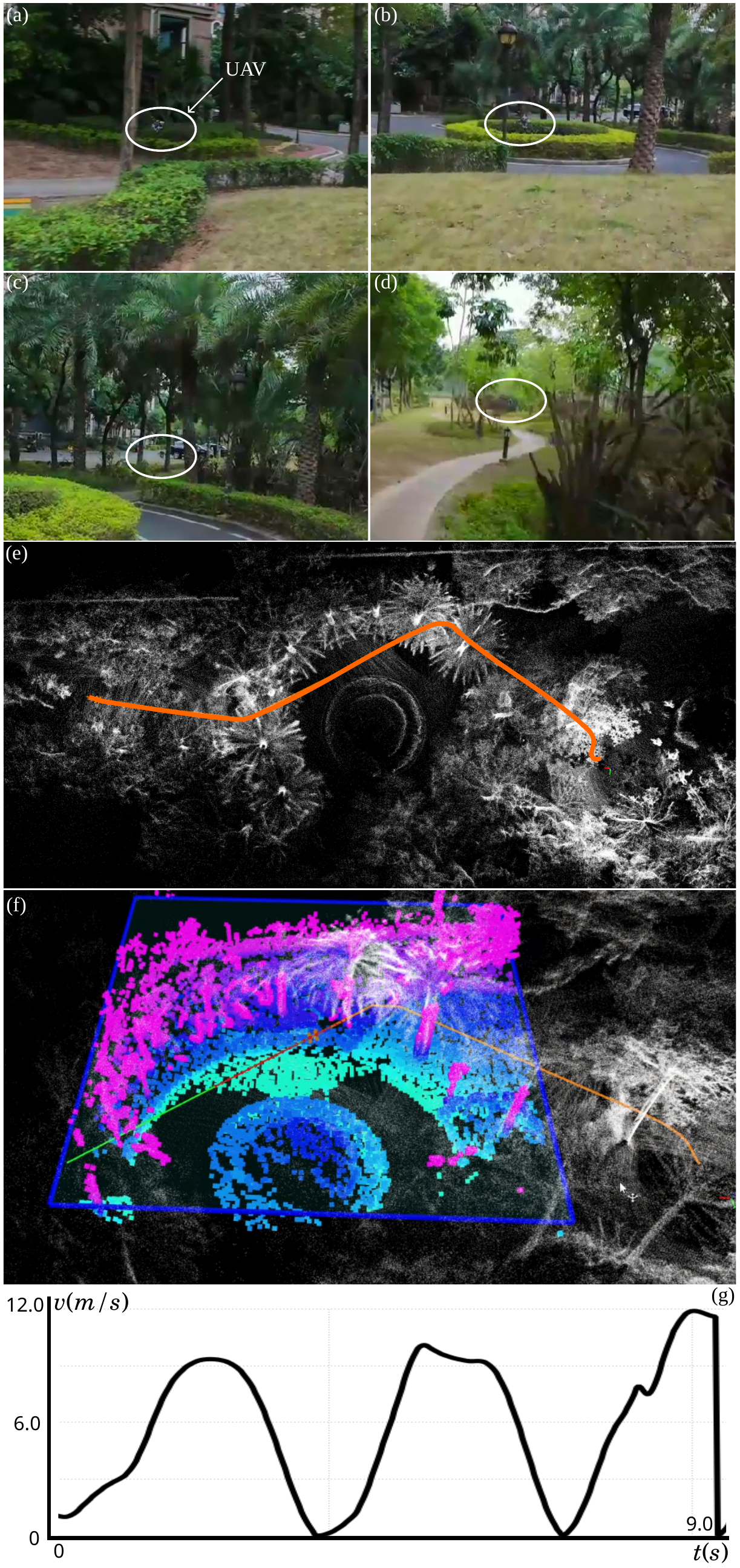}}
\caption{The results of an outdoor high-speed flight test. (a)-(d) are snapshots of the flight process. (b) is the rviz visualization of the flight history. (c) is the rviz visualization of the planning process. (d) is the speed curve of the UAV.}
\label{f: realworld2}
\end{figure}

In the second scenario, the UAV is required to fly across the wild forest with the speed limit $v_{lim} =$ 13.0 m/s. Fig. \ref{f: realworld2} presents the result that the UAV flies through 3 given waypoints in the forest, and Fig. \ref{f: cover} shows a composed image of the UAV navigating to a given target. The UAV can navigate with a maximum speed of over 11.0 m/s and succeeds in multiple tests despite the high density of obstacles and the short duration available for replanning. Repeated experiments show similar patterns and are not shown in this section. More experiment performances can be found in the attached video.

\section{Conclusion}
In this article, we proposed a fast and robust navigation system for high-speed flights of UAVs. An incremental robocentric point cloud map was proposed for efficient collision check and distance and gradient acquisition. An obstacle-aware topological path searching was devised to escape from local minima and explore the solution space more thoroughly, which guides the motion planning to approximate the global optimum. Then, we proposed an adaptive trajectory generation strategy to obtain the spatial-temporal optimal trajectory. The overall system was thoroughly tested in both simulation and real-world scenarios, which showed superior performances of UAVs' high-speed navigation in unknown environments.  

One limitation of our system is that the selection of the best trajectory is only based on the cost value. This strategy is not optimal in all conditions because some demands are not reflected in the cost functions, such as trajectory length. We expect to explore a more thorough trajectory selection method in the future.


%





\ifCLASSOPTIONcaptionsoff
  \newpage
\fi



%

\bibliographystyle{ieeetr}
\bibliography{ref}

%








\end{document}